\documentclass{article}

\PassOptionsToPackage{numbers, compress}{natbib}

\usepackage[final]{neurips_2024}

\usepackage[utf8]{inputenc} 
\usepackage[T1]{fontenc}    
\usepackage{hyperref}       
\usepackage{url}            
\usepackage{booktabs}       
\usepackage{amsfonts}       
\usepackage{nicefrac}       
\usepackage{microtype}      
\usepackage[table]{xcolor}         
\usepackage{graphicx}
\usepackage{amsmath}
\usepackage{amssymb}
\usepackage{bbm}
\usepackage{multicol}
\usepackage{multirow}
\usepackage{algorithm}
\usepackage{algorithmic}
\usepackage{tocbasic}
\usepackage{titletoc}

\definecolor{limegreen}{rgb}{0.2, 0.8, 0.2}
\definecolor{princetonorange}{rgb}{1.0, 0.56, 0.0}
\definecolor{royalpurple}{rgb}{0.47, 0.32, 0.66}
\definecolor{airforceblue}{rgb}{0.36, 0.54, 0.66}
\definecolor{customorange}{HTML}{ff7f0e}
\definecolor{customblue}{HTML}{1f77b4}
\definecolor{customgold}{HTML}{ffcd3f}

\def\MethodName{AdaNCA}
\def\FullMethodName{AdaNCA}
\def\Perception{Interaction}
\def\dynperception{Dynamic Interaction}
\def\Adaptation{Update}

\newcommand{\rev}[1]{\textcolor{black}{#1}}

\usepackage{pifont}
\usepackage{enumitem}
\usepackage{wrapfig}

\newcommand{\cmark}{\textcolor{limegreen}{\ding{51}}}%
\newcommand{\xmark}{\textcolor{red}{\ding{55}}}%

\title{\FullMethodName{}: Neural Cellular Automata as Adaptors for More Robust Vision Transformer}

\author{%
  Yitao Xu \quad Tong Zhang  \quad Sabine Süsstrunk\\
  Image and Visual Representation Lab \\
  École polytechnique fédérale de Lausanne,Lausanne, Switzerland \\
  \texttt{\{yitao.xu,tong.zhang,sabine.susstrunk\}@epfl.ch} \\
}

\begin{document}

\maketitle

\begin{abstract}
    Vision Transformers (ViTs) demonstrate remarkable performance in image classification \rev{through visual-token interaction learning}, particularly when equipped with local information via region attention or convolutions. Although such architectures improve the feature aggregation from different granularities, they often fail to contribute to the robustness of the networks. Neural Cellular Automata (NCA) enables the modeling of global \rev{visual-token} representations through local interactions, as its training strategies and architecture design confer strong generalization ability and robustness against noisy input. In this paper, we propose \textbf{Ada}ptor \textbf{N}eural \textbf{C}ellular \textbf{A}utomata (\FullMethodName{}) for Vision Transformers that uses NCA as plug-and-play adaptors between ViT layers, thus enhancing ViT's performance and robustness against adversarial samples as well as out-of-distribution inputs. To overcome the large computational overhead of standard NCAs, we propose \textit{\dynperception{}} for more efficient interaction learning. Using our analysis of \MethodName{} placement and robustness improvement, we also develop an algorithm for identifying the most effective insertion points for \MethodName{}. With less than a 3\% increase in parameters, \MethodName{} contributes to more than 10\% of absolute improvement in accuracy under adversarial attacks on the ImageNet1K benchmark. Moreover, we demonstrate with extensive evaluations across eight robustness benchmarks and four ViT architectures that \MethodName{}, as a plug-and-play module, consistently improves the robustness of ViTs. 
\end{abstract}

\section{Introduction}
Vision Transformers (ViTs) exhibit impressive performance in image classification, through globally modeling token interactions via self-attention mechanisms~\cite{scale-vit-22b,vit,scale-vit}. Recent works show that integrating local information into ViTs, \textit{e.g.,} using region attention ~\cite{neighbor-attention,swin,swinv2,sasa,halonet,wu2021cvt,focal-attn-local-global} or convolution~\cite{cpvt,coatnet,d2021convit,guo2022cmt,robustfy-attention-tapadl,localvit,rvt,conformer,evolving-attn-conv,conv-head-vit,ceit-conv-in-mlp}, further enhances the ViT's capabilities in image classification. Although advanced local structures contribute to better captures of local information, the robustness of ViTs has not increased. They remain vulnerable to noisy input such as adversarial samples~\cite{cw,autoattack,imagenet-a,pgd,rvt} and out-of-distribution (OOD) inputs~\cite{imagenet-texture,shortcut-learning,imagenet-c,imagenet-r,imagenet-sk}.

Recently, Neural Cellular Automata (NCA) was proposed as a lightweight architecture for modeling local cell interactions \cite{mordvintsev2020growing}, where cells are represented by 1D vectors. To perform downstream tasks, similarly to the idea of token interactions in ViTs, cells in NCA interact with each other by alternating between a convolution-based \textit{\Perception{}} stage and an MLP-based \textit{\Adaptation{}} stage \cite{niklasson2021self-sothtml, meshnca}. The critical difference, however, is that cell interactions in NCA evolve over time by recurrent application of the two stages, whereas ViT computes the token interaction in a single step per layer. During this process, cells dynamically modulate their representations, based on the interactions with their neighbors, and they gradually enlarge their receptive fields. Unlike commonly used convolutional neural networks, NCA maintains resolution during neighborhood expansion. The \rev{recurrent update scheme enables the cells to explore various states}, thus preventing NCA from overfitting and enhancing its generalization ability \cite{mordvintsev2020growing,niklasson2021self-sothtml}. NCA training involves various kinds of stochasticity  \cite{asynchronicity-nca}, which enables the models to generalize to input variability and adapt to unpredictable perturbations. It is the modulation of local information and stochasticity during training that make NCA robust against noisy input \cite{meshnca,dynca,randazzo2020self-classifying,attention-nca}. 
\begin{wrapfigure}{r}{0.5\linewidth}
    \centering
    \includegraphics[width=1.0\linewidth, trim={150 115 150 115},clip]{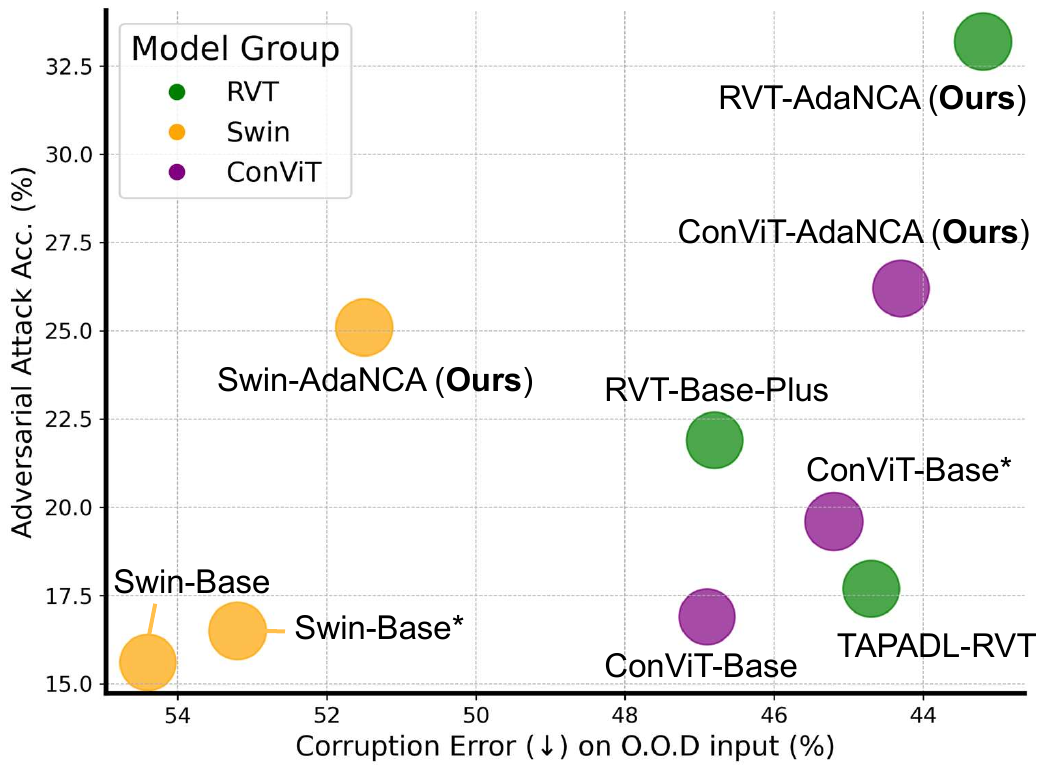}
    \caption{The accuracy under adversarial attacks (APGD-DLR \cite{autoattack}) versus corruption error on out-of-distribution input (ImageNet-C \cite{imagenet-c}) of various ViT models \cite{d2021convit,robustfy-attention-tapadl,swin,rvt}. \MethodName{} improves the robustness of different ViTs against both adversarial attacks and OOD input.
    $\star$: the same model architecture but with more layers.
    }
    \vspace{-10pt}
    \label{fig:teaser}
\end{wrapfigure}

However, the original NCA has substantial computational overhead when operating in high-dimensional space, which is a common scenario in ViTs. This poses a non-trivial challenge when integrating NCA into ViTs. To reduce the dimensionality of interaction results, hence to lower the computational cost, we propose \textit{\dynperception{}} to replace the standard \textit{\Perception{}} stage. In this stage, tokens dynamically modify the interaction strategy, based on the observation of their environment. This adaptation to the variability of the environment contributes to the model robustness. Our modified \textbf{Ada}ptor \textbf{NCA} (\FullMethodName{}), as a plug-and-play module, improves ViTs performances, as illustrated in Figure \ref{fig:teaser}. Adding \MethodName{} to different ViT architectures consistently improves their robustness to both adversarial attacks and OOD input. AdaNCA also improves clean accuracy.

Motivated by empirical observations of the positive relationship between network redundancy and model robustness \cite{imagenet-c}, we develop a dynamic programming algorithm for computing the most effective insert position for \MethodName{} within a ViT, based on our proposed quantification of network redundancy. Our method results in consistent improvements across eight robustness benchmarks and four different baseline ViT architectures. Critically, we demonstrate that the improvements do not originate from the increase in parameters and FLOPS but are attributed to \MethodName{}. Our contributions are as follows:
\begin{itemize}[leftmargin=*]
    \item We propose \FullMethodName{}: It integrates \textbf{N}eural \textbf{C}ellular \textbf{A}utomata into ViTs' middle layers as lightweight \textbf{Ada}ptors for the robustness enhancement of ViTs against adversarial attacks and OOD inputs in image classification. With less than 3\% more parameters, \MethodName{}-extended ViTs can, under certain adversarial attacks, achieve 10\% higher accuracy.
    \item We introduce \textit{\dynperception{}} to replace the \textit{\Perception{}} stage in standard NCA, thus enhancing model robustness and efficiency in terms of parameters and computation.
    \item We propose a method for determining the most effective insert positions of AdaNCA, for maximum robustness improvement. 
\end{itemize}
\begin{figure}
    \centering
    \includegraphics[width=1.0\linewidth, trim={40 150 40 150},clip]{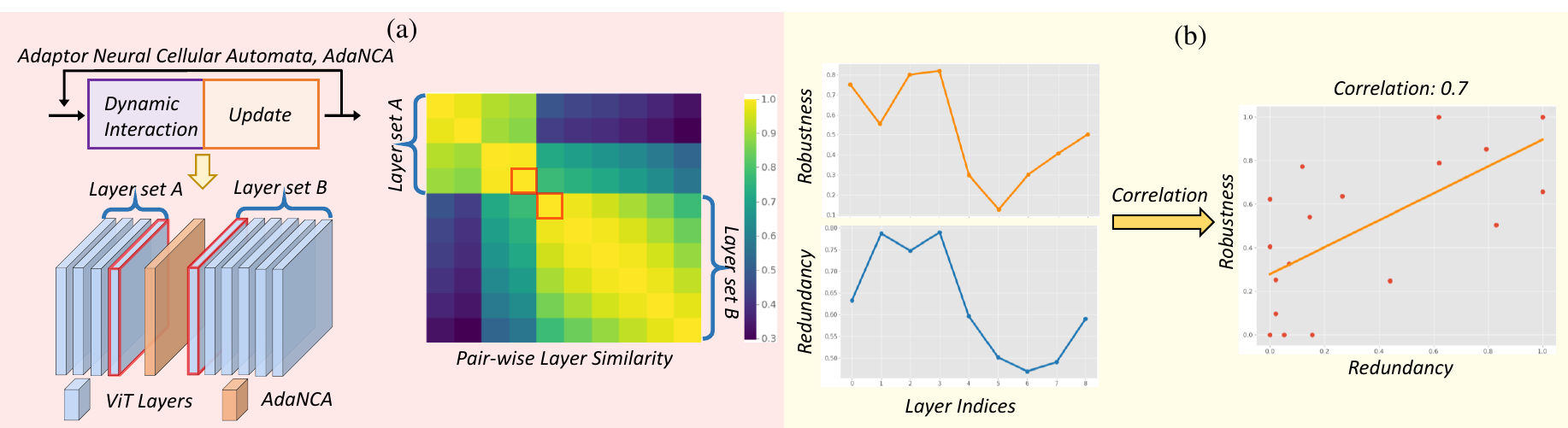}
    \caption{Method overview. (a) To improve model performance and robustness, Neural Cellular Automata (NCA) can be inserted into Vision Transformers (ViTs) as \textbf{Ada}ptors, hence termed \FullMethodName{}. \rev{The details of \MethodName{} are presented in Section~\ref{sec:enca-arch}.} The improvement is maximized when \MethodName{} is inserted between two layer sets that each consists of similar layers. (b) The robustness improvement brought by \MethodName{} is highly correlated with the corresponding network redundancy quantification of the insert position introduced in Section~\ref{sec:insert-pos}. This supports the idea that \MethodName{} should be placed between two sets of redundant layers.}
    \label{fig:overview}
\end{figure}

\section{Related Works}
\subsection{Local structure in Vision Transformers}
Since the proposal of Vision Transformer (ViT) \cite{vit}, a series of works have introduced local structures into ViTs to enhance their performance \cite{cpvt,coatnet,d2021convit, guo2022cmt,robustfy-attention-tapadl,localvit,rvt,conformer,attention-nca,evolving-attn-conv, wu2021cvt,conv-head-vit,focal-attn-local-global,ceit-conv-in-mlp,t2t}. Here, we mention one of the earliest local structure modifications and those relevant to our work.
Stand-Alone Self-Attention (SASA), as introduced by \citet{sasa}, utilizes sliding window self-attention in ViTs. Following this, \citet{swin} develop a non-sliding window attention mechanism that partitions feature maps and computes self-attention, both within and between these partitions; it is termed Shifted Window (Swin) attention. Another method for modeling local information is convolution. \citet{d2021convit} introduce a soft local-inductive bias by using gated positional self-attention, thus fusing self-attention and convolution. 
Despite these advancements in better modeling of local information, few methods lead to a more robust ViT architecture \cite{rvt}. This leaves the models to behave subpar when encountering slightly noisy inputs or distribution shifts.
\subsection{Robust architecture in Vision Transformers}
Researchers have developed various architectural changes for building more robust ViTs against adversarial attacks, such as FGSM \cite{fgsm} or PGD \cite{pgd}, as well as out-of-distribution (OOD) inputs, such as image corruption \cite{imagenet-c}. \citet{fan} propose Full Attention Networks to boost the robustness of ViTs against OOD images. \citet{rvt} first systematically analyze the relationship between different components in a ViT, drawing a positive relationship between convolutional components and the robustness of ViTs against adversarial samples and OOD data. By extending \cite{rvt} and \cite{fan}, \citet{robustfy-attention-tapadl} propose input-dependent average pooling in order to adaptively select different aggregation neighborhoods for different tokens, thus achieving the state-of-the-art robust ViTs in OOD generalization. Different interpretations of the self-attention operation can also lead to more robust architectures \cite{vit-kde,attention-sparse-recurrent}. However, the methods that introduce additional architectures \cite{robustfy-attention-tapadl,vit-kde,attention-sparse-recurrent,fan} are either implemented on ViTs with limited size or focus on non-adversarial robustness. On the contrary, our method introduces NCA as lightweight plug-and-play adaptors into base-level ViTs, thus enhancing their clean accuracy and robustness against both adversarial samples and OOD inputs. 
\subsection{Neural Cellular Automata}
\rev{\citet{nca-cnn-2} demonstrates that CA can be represented by convolutional neural networks. By extending \cite{nca-cnn-2}}, \citet{mordvintsev2020growing} propose NCA in order to mimic the biological cell interactions and model morphogenesis. Following this idea, several works apply NCA in computer vision, including texture synthesis \cite{mordvintsev2021mu-micronca,niklasson2021self-sothtml,meshnca,dynca,noisenca}, image generation \cite{freq-time-diffusion-nca,gan-nca,vnca,attention-nca}, and image segmentation \cite{mednca,nca-imageseg}. \citet{randazzo2020self-classifying} propose applying NCA for modeling collective intelligence on image classification tasks, but the limitation to binary images restricts its practical application. \citet{attention-nca} first establish the connection between ViT and NCA via recurrent local attention. \rev{It leads to a more robust model for handling image corruptions in image inpainting tasks.} However, their application is limited to image impainting on small datasets such as MNIST \cite{mnist} and CIFAR10 \cite{cifar10}. \rev{The NCA in \cite{attention-nca} attempts to emulate a ViT whereas our approach distinguishes itself by not doing so.} We first applies NCA in image classification on ImageNet1K with base-level ViT models. Moreover, we propose the new \textit{\dynperception{}} for efficient cell interaction modeling, reducing the computational overhead, and for enhancing the model performance.

\section{Method}
The overview of our method is shown in Figure~\ref{fig:overview}. In this section, we first review the NCA model and ViT architecture. In Section~\ref{sec:preliminary}, we establish the connection between NCA and ViT, in terms of token interaction modeling. We then present the design of our \MethodName{} in Section~\ref{sec:enca-arch}. We insert \MethodName{} into the middle layers of ViT to improve its robustness.  We introduce the relationship between the insert position of \MethodName{} and the relative improvements of model robustness in Section~\ref{sec:insert-pos}. This relationship leads to the algorithm for deciding the most effective placement for \MethodName{}.

\subsection{Preliminaries}
\label{sec:preliminary}
\paragraph{Vision Transformers}
Vision Transformers (ViTs) operate on token maps $\mathbf{X} \in \mathbb{R}^{N \times C}$, where the number of tokens is $N$ and each token is represented by a $C$-dimensional vector. ViTs learn the interaction between these tokens via self-attention \cite{nlp-transformer} and compute the interaction result $\mathbf{X}_{attn}$, as described in Equation~\ref{eq:self-attn}.
\begin{equation}
    \mathbf{X}_{attn} = \sigma \left (\frac{\mathbf{Q}\mathbf{K}^\top}{\sqrt{C}} \right ) \mathbf{V}.
    \label{eq:self-attn}
\end{equation}
$\mathbf{Q},\mathbf{K},\mathbf{V}$ stand for query, key, and value, respectively. They are deduced from different linear projections of the input, i.e., $\mathbf{Q}=\mathbf{X}\mathbf{W}_\mathbf{Q},\mathbf{K}=\mathbf{X}\mathbf{W}_\mathbf{K},\mathbf{V}=\mathbf{X}\mathbf{W}_\mathbf{V}$, where $\mathbf{W}_\mathbf{Q},\mathbf{W}_\mathbf{K},\mathbf{W}_\mathbf{V} \in \mathbb{R}^{C \times D}$. $D$ is the hidden dimensionality in self-attention. $\sigma$ is Softmax. After self-attention, tokens are fed into a Multilayer Perceptron (MLP) to obtain the updated representations $\mathbf{X}_{out}$:
\begin{equation}
    \mathbf{X}_{out} = f_{\theta} (\mathbf{X}_{attn}).
\end{equation}
$f$ is the MLP and $\theta$ stands for its parameters. The self-attention and MLP form a single ViT block, and a ViT model can be built via stacking ViT blocks. 

\paragraph{Neural Cellular Automata}
NCA aims at modeling cell interactions. In the 2D domain, cells live on a 2D grid with size $H \times W$. Each cell is represented by a vector with dimensionality $C$. All cells collectively define the cell states $\mathbf{S} \in \mathbb{R}^{H \times W \times C}$. In a single step of NCA, to generate the interaction output $\mathbf{S}_{\mathcal{I}}$, cells first interact with their neighborhoods for an information exchange in the \textit{\Perception{}} stage \cite{meshnca}; the interaction is typically instantiated via depth-wise convolutions \cite{mordvintsev2020growing,niklasson2021self-sothtml, dynca}:
\begin{equation}
    \mathbf{S}_{\mathcal{I}} = (\mathbf{S} \circledast [\mathcal{C}_1,\mathcal{C}_2,...,\mathcal{C}_\mathcal{M}])_{\oplus}.
    \label{eq:nca-perception}
\end{equation}
$\mathcal{C}_i$ is the $i$th convolutional kernel, and $\mathcal{M}$ denotes the total number of kernels. \rev{`$\circledast$' denotes depth-wise convolution.} The kernels can either be fixed \cite{meshnca,dynca} or learnable \cite{mordvintsev2020growing,niklasson2021self-sothtml}. The results of all kernels are concatenated channel-wise in the $\oplus$ operation. $\mathbf{S}_{\mathcal{I}}$ is then passed to an MLP in the \textit{\Adaptation{}} stage \cite{meshnca}:
\begin{equation}
    \mathbf{S}_{out} = f_\theta(\mathbf{S}_{\mathcal{I}}) \odot \mathbf{M}
    \label{eq:nca-adaptation}.
\end{equation}
\rev{$\mathbf{S}_{out}$ is then used to update the cell states in a residual scheme.} $f$ is the MLP, and $\theta$ stands for its parameters. Typically, NCA uses the simplest MLP, with two linear layers and one activation between them. $\mathbf{M} \in \mathbb{R}^{H \times W \times C}$, sampled from $Bernoulli(p)$, is a random binary mask to introduce stochasticity in NCA; it ensures asynchronicity during the cell updates \cite{asynchronicity-nca}. $\odot$ is point-wise multiplication. NCA learns an underlying dynamic that governs the cell behaviors \cite{noisenca}, as depicted by the stochastic differential equation (SDE) in Equation~\ref{eq:sde}:
\begin{equation}
    \frac{\partial \mathbf{S}}{\partial t} = \mathcal{F}_{\Theta} (\mathbf{S}) \odot \mathbf{M}
    \label{eq:sde} = f_\theta \left [ (\mathbf{S} \circledast [\mathcal{C}_1,\mathcal{C}_2,...,\mathcal{C}_\mathcal{M}])_{\oplus} \right ] \odot \mathbf{M}.
\end{equation}
$\mathcal{F}$ represents operations in \textit{\Perception{}} as well as \textit{\Adaptation{}} stages. $\Theta$ is the set containing trainable parameters in the two stages. Discretizing the SDE with $\Delta t = 1$ naturally results in a recurrent residual update scheme:
\begin{equation}
    \mathbf{S}^{t+1} = \mathbf{S}^t+\mathcal{F}_{\Theta} (\mathbf{S}^t) \odot \mathbf{M}.
    \label{eq:sde-discrete}
\end{equation}
\rev{After $t=T$ steps, the cell states $S^T$ is extracted to accomplish certain downstream tasks. The traditional NCA involves several other specific designs though, in our case, it is impractical to adapt them. We provide a discussion on this topic in Appendix~\ref{sec:diff-nca-adanca}.}

\subsubsection{Connecting NCA and ViT}
\label{sec:connect-nca-vit}
Both NCA and ViT learn interactions between a set of elements, i.e., tokens in ViT and cells in NCA. Hereafter, we refer to a cell in NCA as a token, aligning it with the concept in ViTs. The asynchronicity \cite{asynchronicity-nca} introduced by the random mask $\mathbf{M}$ can be regarded as a cell-wise stochastic depth \cite{stochastic-depth}, a more fine-grained version of the sample-wise stochastic depth. In previous NCA works, stochasticity is maintained during testing \cite{niklasson2021self-sothtml}. \rev{Such a scheme is problematic in our case because (1) test-time stochasticity produces obfuscated gradients \cite{obfuscated-grad}, leading to the circumvention of adversarial attacks, and (2) the model can output different results given the same inputs.} To this end, we adopt the strategy in dropout-like techniques \cite{stochastic-depth,dropout}, which compensates activation values during training. Given $\mathbf{M} \sim Bernoulli(p)$, the evolution of NCA in \FullMethodName{} is defined as:
\begin{equation}
    \textbf{Train}: \mathbf{S}^{t+1} = \mathbf{S}^t+\frac{\mathcal{F}_{\Theta} (\mathbf{S}^t)}{p} \odot \mathbf{M}; \quad \textbf{Test}: \mathbf{S}^{t+1} = \mathbf{S}^t+\mathcal{F}_{\Theta} (\mathbf{S}^t).
    \label{eq:nca-vit-train}
\end{equation}
\rev{We discuss the necessity of such a scheme in Section~\ref{sec:dropout-stoch-abl} in the Appendix.} Furthermore, NCA typically \rev{outputs} the cell states at a random time step $T$, resulting in random update steps for all cells. Such randomness ensures the stability of NCA across various time steps \cite{mordvintsev2020growing}. \rev{Finally, the recurrent steps of NCA during a single training epoch enable the exploration of a wide range of cell states. In the early stage of training, the model is not adequately trained hence serves as a source of noise to itself through the recurrence. With all these components,} the trained model can effectively handle the variability and unpredictability of the input thus be robust against noisy input. Our ablation studies in Section~\ref{sec:ablation} demonstrate the effectiveness of these strategies in enhancing the model performance.

\definecolor{custompurple}{HTML}{7030A0}
\definecolor{customgreen}{HTML}{75B446}
\definecolor{customorange1}{HTML}{F08746}

\subsection{\MethodName{} architecture}
\label{sec:enca-arch}
\begin{figure*} 
    \centering
    \includegraphics[width=1.0\linewidth,trim={80 55 80 50},clip]{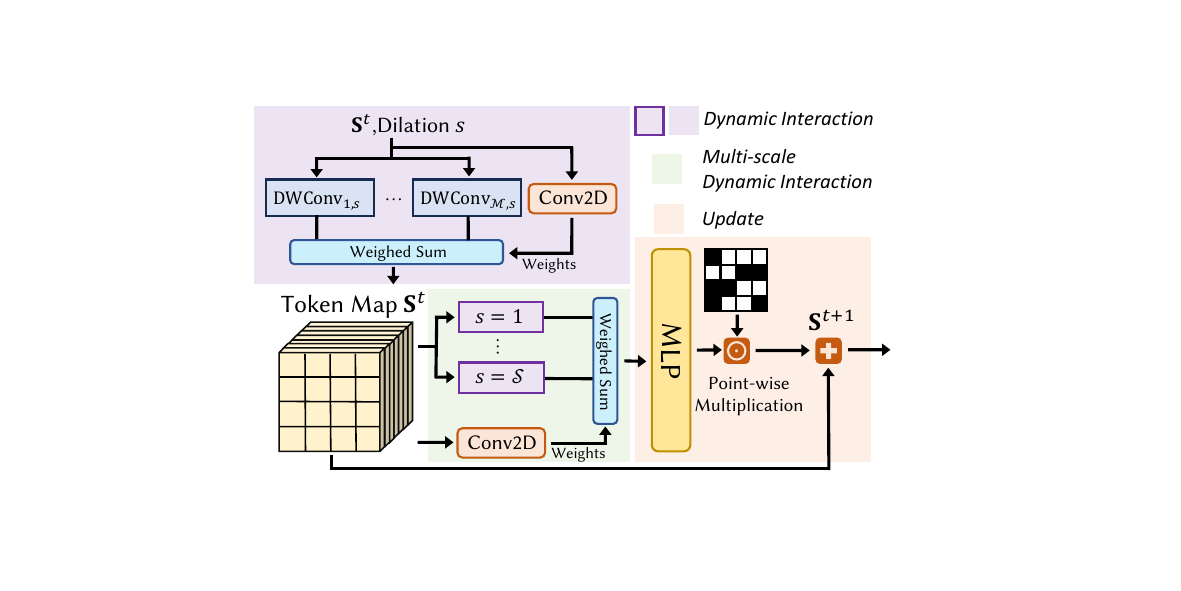}
    \caption{Overview of \MethodName{} architecture. Instead of concatenating the interaction results generated by the depth-wise convolutions, our \textcolor{custompurple}{\textit{\dynperception{}}} conducts a point-wise weighted sum on them to improve the efficiency and enhance the performance. The weights are obtained based on the token states so that each token can dynamically adjust, according to the inputs, the interaction strategy. The \textcolor{customgreen}{\textit{Multi-scale \dynperception{}}} aggregates the results from \textcolor{custompurple}{\textit{\dynperception{}}} , where the convolutions have different dilation rates. Then, to finish one step of evolution, the output is fed into the \textcolor{customorange1}{\textit{\Adaptation{}}} stage .}
    \vspace{-15pt}
    \label{fig:adanca}
\end{figure*}
The architecture of \MethodName{} is shown in Figure~\ref{fig:adanca}. It shares a similar update scheme with the standard NCA, as described in Equation~\ref{eq:nca-vit-train}; but, it is more computationally efficient due to the proposed \textit{\dynperception{}} stage. All convolutional kernels for token interaction are trainable. In the following paragraphs, we first present the design of the \textit{\dynperception{}} stage and then introduce a way for more efficient token interaction by using multi-scale \textit{\dynperception{}}. 
\subsubsection{\dynperception{}}
The \textit{\Perception{}} stage in the original NCA performs a channel-wise concatenation of the \rev{$\mathcal{M}$ output tensors} from different depth-wise convolutions. Whereas, our \textit{\dynperception{}} computes a weighted sum of those results.
Specifically, a weight computation \rev{network $\mathcal{W}_{\mathcal{I}}$ takes the token map $\mathbf{S} \in \mathbb{R}^{H \times W \times C}$} as input and outputs per-token scalar weights $\mathbf{W}_{\mathcal{I}m} \in \mathbb{R}^{H \times W \times 1}$ \rev{for each of the $\mathcal{M}$ kernels.} 
We modify Equation~\ref{eq:nca-perception} to Equation~\ref{eq:dynamic-perception}:
\begin{equation}
    \mathbf{S}_{D\mathcal{I}} = \sum (\mathbf{S}*\mathcal{W}_{\mathcal{I}}) (\mathbf{S} \circledast [\mathcal{C}_1,\mathcal{C}_2,...,\mathcal{C}_\mathcal{M}])
    = \sum (\mathbf{S}*\mathcal{W}_{\mathcal{I}}) \mathbf{S}_{m}
    = \sum_{m=1}^{\mathcal{M}} \left ( \sum_{c=1}^{C} \mathbf{W}_{\mathcal{I}m} \odot \mathbf{S}_{mc} \right ),
    \label{eq:dynamic-perception}
\end{equation}
where $ \mathbf{S}_{mc} \in \mathbb{R}^{H \times W \times 1}, \mathbf{S}_{D\mathcal{I}}, \mathbf{S}_{m} \in \mathbb{R}^{H \times W \times C}$.\rev{`$*$' denotes the convolution. Recall that `$\circledast$' is the depth-wise convolution.} We instantiate the weight computation module $\mathcal{W}_\mathcal{I}$ by using a two-layer convolutional network. The first layer transforms the dimensionality from $C$ to $\mathcal{M}$, and the second layer computes the actual weights, thus producing $\mathcal{M}$ scalars for each token. Both layers use $3 \times 3$ convolutions to factor in information from both the token and its neighbors. To stabilize training, we add a batch normalization between the two convolutions. Our design of the weight computation network coincides with the one in \cite{robustfy-attention-tapadl}. Although, our focus is on extracting various information from the same neighborhoods rather than on aggregating data from different neighborhoods.

\subsubsection{Multi-scale \dynperception{}}
Inspired by \cite{dynca}, which uses multi-scale token \textit{\Perception{}} to facilitate long-range token communication, we propose multi-scale \textit{\dynperception{}}. Concretely, all convolutions in Equation~\ref{eq:dynamic-perception} now have one more degree of freedom in dilation. Dilation $s$ represents the current operating scale being $s$, and $s \in \{1,2,...,\mathcal{S}\}$. Hence, the original \textit{\dynperception{}} is a special case where $\mathcal{S}=1$. To increase the feature expressivity, we perform a weighted sum on the outputs of all scales, where the per-token weights $\mathbf{W}_{Ms}$ are generated by a network $\mathcal{W}_{M}$ as described in Equation~\ref{eq:multi-scale-interaction}.
\begin{equation}
    \mathbf{S}_{MD\mathcal{I}} = \sum (\mathbf{S}*\mathcal{W}_{M}) \mathbf{S}_{D\mathcal{I}} 
    = \sum (\mathbf{S}*\mathcal{W}_{M}) \mathbf{S}_{D\mathcal{I}}
    = \sum_{s=1}^{\mathcal{S}} \left ( \sum_{c=1}^{C} \mathbf{W}_{Ms} \odot \mathbf{S}_{D\mathcal{I}sc} \right ),
    \label{eq:multi-scale-interaction}
\end{equation}
where $\mathbf{W}_{Ms},\mathbf{S}_{D\mathcal{I}sc} \in \mathbb{R}^{H \times W \times 1}, \mathbf{S}_{MD\mathcal{I}},\mathbf{S}_{D\mathcal{I}s} \in \mathbb{R}^{H \times W \times C}$. \rev{The $\mathcal{W}_{\mathcal{I}}$ in Equation~\ref{eq:dynamic-perception} is shared across all scales.} The weight computation network $\mathcal{W}_{M}$ \rev{mirrors that of $\mathcal{W}_{\mathcal{I}}$}.

\subsection{Insert positions of \MethodName{}}
\label{sec:insert-pos}
Given a ViT and an \MethodName{}, to maximize the robustness improvements, we need to determine where to insert \MethodName{}. To this end, we first establish the correlation between the placement of \MethodName{} and the robustness enhancement it brings. Motivated by the fact that the network redundancy contributes to the model robustness \cite{imagenet-c}, we hypothesize that the effect of \MethodName{} should correlate to the layer redundancy corresponding to the insert position. To quantify the redundancy, we propose the \textbf{Set Cohesion Index} $\kappa$. Given a trained model with $L$ layers and two layer indices $i,j \in \{1,2,...,L\}$ where $i<j$, $\kappa(i,j)$ is defined in Equation~\ref{eq:stage-cohesion-index}.
\begin{equation}
    \begin{aligned}
        \kappa(i,j) = \frac{1}{(j-i+1)^2} \sum_{m,n \in \left [i,j\right ] } Sim(m,n) & - \frac{\mathbbm{1}_{i>1}}{(i-1)(j-i+1)} \sum_{m_1 \in [1,i-1], n \in \left [i,j\right ]} Sim(m_1,n) \\
        &  - \frac{\mathbbm{1}_{j<L}}{(L-j)(j-i+1)} \sum_{m \in \left [i,j\right ],n_1 \in [j+1,L]} Sim(m,n_1)
    \end{aligned}
    \label{eq:stage-cohesion-index}
\end{equation}
$\mathbbm{1}$ stands for the indicator function. $Sim(m,n)$ is the function for quantifying the output similarity between layer $m$ and $n$. We choose Centered Kernel Alignment (CKA) \cite{cka}, a common metric for measuring layer similarities inside or between neural networks \cite{convergent-feature-different-network}. A higher $\kappa$ stands for a more cohesive layer set defined by layers from $i$ to $j$. Inserting \MethodName{} after layer $i$ would partition the network into two layer sets, and we can compute the sum of $\kappa$ of the layers before and after $i$, \textit{i.e.}, $\mathcal{K}(i)=\kappa(1,i)+\kappa(i+1,L)$. This serves as a quantification of the network redundancy that corresponds to position $i$. We assume that \MethodName{} will not change the layer similarity structure because it is too small compared to a single layer in all ViTs.

In addition to the quantification of the network redundancy, the robustness improvement, brought by \MethodName{}, is quantified  using the relative increase in the attack failure rate of the \MethodName{}-inserted models and the corresponding baseline. Specifically, if a model can achieve $\alpha$ clean test accuracy as well as $\alpha^\prime$ accuracy under adversarial attacks, the attack failure rate is $\beta=\frac{\alpha^\prime}{\alpha}$. The robustness improvement $\gamma$ is then defined as $\gamma=\frac{\beta_{\MethodName{}}-\beta_{base}}{\beta_{base}}$. In our experiments, we find that $\gamma$ is significantly correlated with the network redundancy $\mathcal{K}$ (\rev{Pearson correlation} $r=0.6938, p<0.001$). We refer readers to Appendix~\ref{sec:insert-pos-supp} for details of the experiments. The results validate our hypothesis and indicate that \MethodName{} should be inserted into the position that can maximize network redundancy. We develop a dynamic programming algorithm to find these positions and refer readers to Appendix~\ref{sec:dynamic-programming} for the details. 

\section{Experiments}
\label{sec:exp}
We use four ViT models as the baseline: Swin-base (Swin-B) \cite{swin}, FAN-Base-Hybrid (FAN-B) \cite{fan}, RVT-Base-Plus (RVT-B) \cite{rvt}, and ConViT-Base (ConViT-B) \cite{d2021convit}. They include a hierarchical model (Swin), a convolution-attention hybrid model (FAN), and two regular models in which all layers share the same structure (RVT and ConViT). RVT is specifically built to be a robust model, whereas ConViT is not. All four models are equipped with different kinds of local structures. We use two SOTA models in terms of robustness against out-of-distribution (OOD) data, TAPADL-RVT and TAPADL-FAN \cite{robustfy-attention-tapadl} for comparison.
\rev{Note that the SOTA method involves training with an additional loss (ADL) and that, for completeness,  we keep the results from the TAPADL models trained with such a loss. However, as our focus is on the effect of architectural changes, we do not incorporate the ADL loss in training the \MethodName{}-enhanced ViTs.} We follow the training scheme for each model, respectively, to train the \MethodName{}-equipped model from scratch on ImageNet1K. We conduct the analysis presented in Section~\ref{sec:insert-pos} to decide the optimal position to insert multiple \MethodName{} modules, in which the ImageNet1K pre-trained weights for analysis are obtained from the PyTorch Image Models library \cite{timm}. To balance between the computational cost and robustness improvement, we limit the number of \MethodName{} to two or three, depending on the model architecture. The recurrent time step of all \MethodName{} is chosen from $\{[2,2],[2,4],[3,5]\}$, and we follow the design principle of ViT; it is to put more computation in the middle or high layers \cite{swin}. \textbf{We fix the random steps during testing to a single integer chosen from the range} to achieve precise results and ensure non-stochasticity during adversarial attacks. All the activation functions used in the MLP in \MethodName{} are GELU \cite{gelu}, and the input, hidden, and output dimensionalities of the MLP are all the same. We refer readers to Appendix~\ref{sec:train-detail} for the details of the training and insert scheme of \MethodName{}. All of our experiments are performed on four Nvidia A100 GPUs.


\begin{table*}[]
\caption{The performance of \MethodName{}-enhanced ViTs and their corresponding baselines. \rev{We report the mCE for IM-C (lower is better) and accuracy for other benchmarks.} \MethodName{} consistently improves the clean accuracy as well as the robustness to adversarial and out-of-distribution inputs of the baseline models. Note that our models also outperform the larger baselines ($^\star$ sign), indicating that the performance improvement does not merely originate from the increase in the number of parameters or FLOPS. The TAPADL method \cite{robustfy-attention-tapadl} can lead to more vulnerable models compared to the baselines (green numbers). \rev{\textbf{Bold} indicates the best model.} $^\dagger$: the test is conducted on models pre-trained on ImageNet22K \cite{rvt}, see Appendix~\ref{sec:supp-notes-on-result}.\\}
\label{tab:main-result}
\resizebox{\linewidth}{!}{
\begin{tabular}{c|cc|c|ccccc|ccc}
\toprule
\multirow{2}{*}{\textbf{Model}} & Params & FLOPS & \textbf{ImageNet} & \multicolumn{5}{c|}{\textbf{Adversarial Inputs}} & \multicolumn{3}{c}{\textbf{OOD inputs}}\\
& (M) & (G) & Clean Acc. & PGD \cite{pgd} & CW \cite{cw} & APGD-DLR \cite{autoattack} & APGD-CE \cite{autoattack} & IM-A \cite{imagenet-a} & IM-C ($\downarrow$) \cite{imagenet-c} & IM-R \cite{imagenet-r} & IM-SK \cite{imagenet-sk} \\
\midrule
RVT-B \cite{rvt} & \textbf{88.5} & \textbf{17.7} & 82.7 & 29.9 & 21.5$^\dagger$ & 21.9$^\dagger$ & 31.4$^\dagger$ & 28.5 & 46.8 & 48.7 & 36.0 \\
TAPADL-RVT \cite{robustfy-attention-tapadl} & 89.4 & 17.9 & 83.1 & \textcolor{limegreen}{27.6} & \textcolor{limegreen}{19.3} & \textcolor{limegreen}{17.7} & \textcolor{limegreen}{26.8} & \rev{\textbf{32.7}} & 44.7 & 50.2 & 38.6 \\
\rowcolor{gray!20} \textit{RVT-B-\MethodName{}} & 91.0 & 19.0 & \textbf{83.3} & \textbf{36.7} & \textbf{30.2} & \textbf{33.2} & \textbf{36.2} & 31.9 & \textbf{43.2} & \textbf{51.7} & \textbf{39.0} \\
\midrule
FAN-B \cite{fan} & \textbf{50.4} & \textbf{11.7} & 83.9 & 15.0 & 7.6 & 10.4 & 13.1 & 39.6 & 46.1 & 52.7 & 40.8 \\
TAPADL-FAN \cite{robustfy-attention-tapadl} & 50.7 & 11.8 & \textbf{84.3} & 18.6 & 9.2 & 13.5 & 16.9 & \rev{42.3} & \textbf{43.7} & \rev{\textbf{54.6}} & \textcolor{limegreen}{40.7} \\
\rowcolor{gray!20} \textit{FAN-B-\MethodName{}} & 51.7 & 12.4 & 84.1 & \textbf{20.3} & \textbf{10.6} & \textbf{14.1} & \textbf{19.1} & \textbf{42.9} & 44.7 & 53.4 & \textbf{41.0} \\
\midrule
Swin-B \cite{swin} & \textbf{87.8} & \textbf{15.4} & 83.4 & 21.3 & 13.4 & 15.6 & 23.1 & 35.8 & 54.3 & 46.6 & 32.4 \\
Swin-B$^\star$ \cite{swin} & 94.1 & 16.7 & 83.3 & 22.8 & 14.6 & 15.9 & 23.8 & 35.2 & 53.2 & 46.9 & 33.7 \\
\rowcolor{gray!20} \textit{Swin-B-\MethodName{}} & 90.7 & 16.3 & \textbf{83.7} & \textbf{24.1} & \textbf{20.5} & \textbf{25.1} & \textbf{24.8} & \textbf{36.0} & \textbf{51.5} & \textbf{48.2} & \textbf{35.5} \\
\midrule
ConViT-B \cite{d2021convit} & \textbf{86.5} & \textbf{17.7} & \textbf{82.4} & 21.2 & 8.9 & 16.9 & 20.3 & 29.0 & 46.9 & 48.4 & 35.7 \\
ConViT-B$^\star$ \cite{d2021convit} & 93.6 & 19.2 & 82.7 & 24.1 & 10.0 & 20.5 & 23.9 & 30.1 & 45.2 & 49.9 & 37.8 \\
\rowcolor{gray!20} \textit{ConViT-B-\MethodName{}} & 89.0 & 19.0 & \textbf{83.2} & \textbf{29.2} & \textbf{20.1} & \textbf{26.3} & \textbf{28.4} & \textbf{33.0} & \textbf{44.3} & \textbf{51.1} & \textbf{39.1} \\
\bottomrule
\end{tabular}
}
\vspace{-15pt}
\end{table*}

\begin{table*}[]
\caption{Comparisons of corruption error on each corruption type of ImageNet-C. \MethodName{} consistently improves the performance of the baseline model in all categories (Swin-B), and can achieve better results in most categories compared to the SOTA method (TAPADL-RVT). \rev{\textbf{Bold} indicates the best model. Lower is better.}\\}
\label{tab:im-c-more}
\resizebox{\linewidth}{!}{
\begin{tabular}{c|c|ccc|cccc|cccc|cccc}
\toprule
\multirow{2}{*}{Model}  &  \multirow{2}{*}{mCE} & \multicolumn{3}{c|}{\textbf{Noise}}  &  \multicolumn{4}{c|}{\textbf{Blur}} & \multicolumn{4}{c|}{\textbf{Weather}} & \multicolumn{4}{c}{\textbf{Digital}} \\
 & &                                        Gauss & Shot & Impulse & Defocus & Glass & Motion & Zoom & Snow & Frost & Fog  & Bright & Contrast & Elastic & Pixelate & JPEG \\
\midrule
Swin-B \cite{swin} & 54.3 &                 43.8    & 45.7 & 43.8    & 61.6    & 73.2  & 55.5   & 66.5 & 50.0 & 47.4  & 47.9 & 42.1       & 38.7     & 68.8    & 66.4     & 62.6 \\
\rowcolor{gray!20} \textit{Swin-B-\MethodName{}}      &  \textbf{51.5} &  \textbf{39.5} & \textbf{40.9} & \textbf{39.8} & \textbf{59.5} & \textbf{69.5} & \textbf{55.0} & \textbf{65.6} & \textbf{49.4} & \textbf{47.1} & \textbf{40.8} & \textbf{40.6} & \textbf{37.0} & \textbf{66.3} & \textbf{59.9}  & \textbf{61.1}  \\ 
\midrule
TAPADL-RVT \cite{robustfy-attention-tapadl} & 44.7 & 34.6    & 36.1 & 34.0    & 53.9    & 62.3  & 52.2   & \textbf{60.1} & \textbf{41.9} & 44.9  & 35.8 & 37.4       & 31.9     & 57.1    & 41.1     & 47.2 \\
\rowcolor{gray!20} \textit{RVT-B-\MethodName{}}      &  \textbf{43.2} &  \textbf{33.9} & \textbf{36.1} & \textbf{33.9} & \textbf{52.7} & \textbf{59.1} & \textbf{47.8} & 60.5 & 42.6 & \textbf{39.0} & \textbf{33.9} & \textbf{36.8} & \textbf{31.7} & \textbf{53.4} & \textbf{40.0}  & \textbf{46.2}  \\           
\bottomrule
\end{tabular}
}
\vspace{-10pt}
\end{table*}
\subsection{Results on Image Classification}
\label{sec:main-results}
We test all models on the ImageNet1K validation set for clean accuracy. For the adversarial robustness evaluation, we choose common adversarial attack methods PGD \cite{pgd}, CW \cite{cw}, APGD-DLR \cite{autoattack}, and APGD-CE \cite{autoattack}. Moreover, we also include the natural adversarial examples in ImageNet-A (IM-A) \cite{imagenet-a}. For the PGD attack, we align with the settings used in \cite{rvt}: max magnitude $\epsilon = 1$, step size $\alpha=0.5$, steps $t=5$. We refer readers to the Appendix~\ref{sec:supp-adv-detail} for details of the other attacks. For testing the OOD generalization, we use ImageNet-C (IM-C) \cite{imagenet-c}, ImageNet-R (IM-R) \cite{imagenet-r}, and ImageNet-Sketch (IM-SK) \cite{imagenet-sk}. We report the mean corruption error (mCE) on ImageNet-C and accuracy on all other kinds of robustness benchmarks in Table~\ref{tab:main-result}. Our results highlight that \MethodName{}-enhanced ViTs consistently outperform corresponding baselines in various robustness tests as well as in terms of clean accuracy. Importantly, the enlarged baseline models ($\star$ sign) do not bring comparable improvements to \MethodName{}, suggesting that the enhancements do not merely stem from the increase in computational budgets. However, the existing method \cite{robustfy-attention-tapadl} that introduces local structure into ViTs can potentially undermine the adversarial robustness of the baselines. 
In Table~\ref{tab:im-c-more}, we conduct an in-depth study on the corruption errors of the different categories of common corruptions in ImageNet-C. 
The results show that \MethodName{} enhances robustness without the trade-off seen in methods that ignore texture information \cite{imagenet-texture,shape-bias-not-good}. While these methods may improve mCE for non-Blur noise, they often worsen mCE for Blur noise \cite{shape-bias-not-good}. In contrast, \MethodName{} consistently improves robustness across most categories. We refer readers to Appendix~\ref{sec:im1k-supp} for more results.


\subsection{Layer similarity structure}
\begin{figure}
    \centering
    \begin{minipage}{0.49\textwidth}
        \centering
        \includegraphics[width=\textwidth,trim={60 120 60 120},clip]{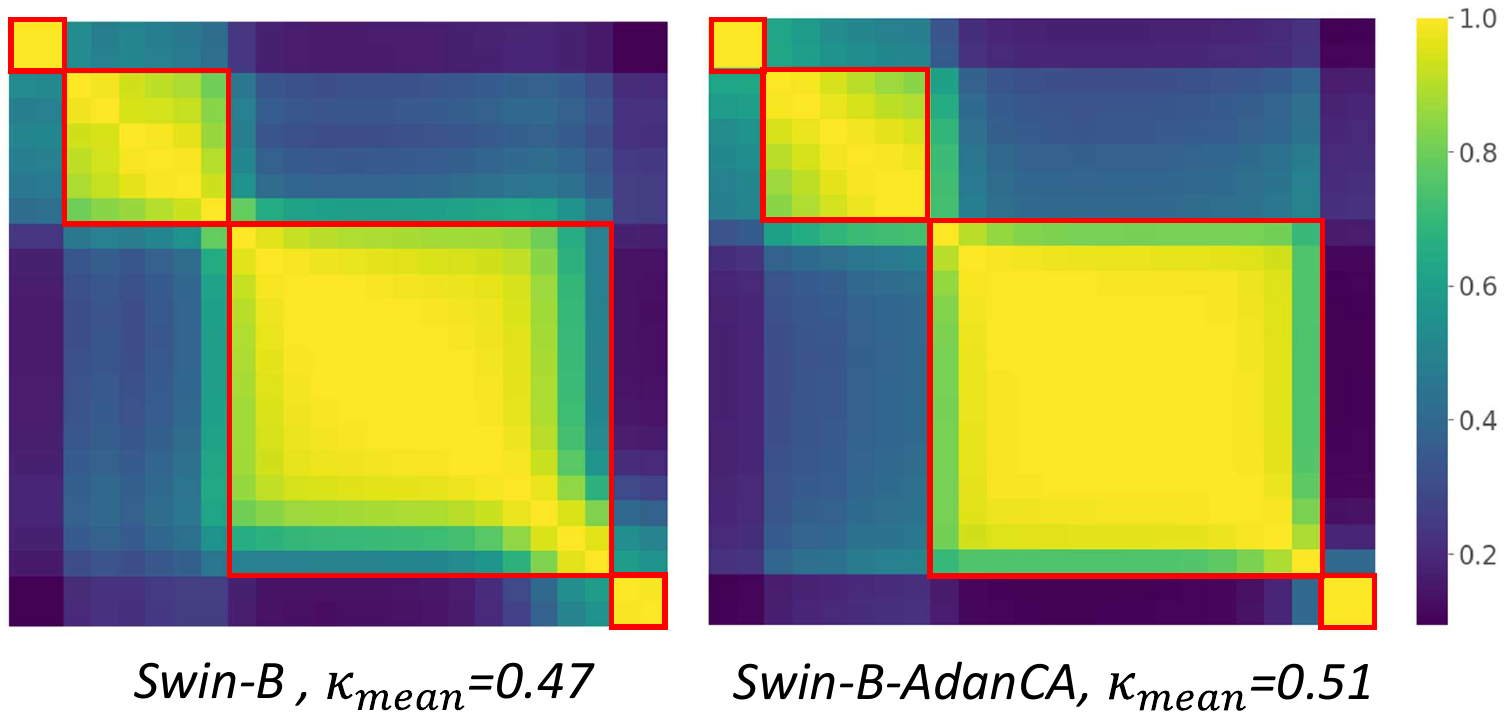}
        \caption{Pair-wise layer similarities. Layer sets are marked in \textcolor{red}{red boxes}. Swin-B-AdanCA has a clearer stage partition, \rev{which might be attributed to \MethodName{} acting as an information transmitter between different layer sets.}}
        \label{fig:layer-sim-swin}
    \end{minipage}\hfill
    \begin{minipage}{0.49\textwidth}
        \centering
        \includegraphics[width=\textwidth,trim={145 100 145 140},clip]{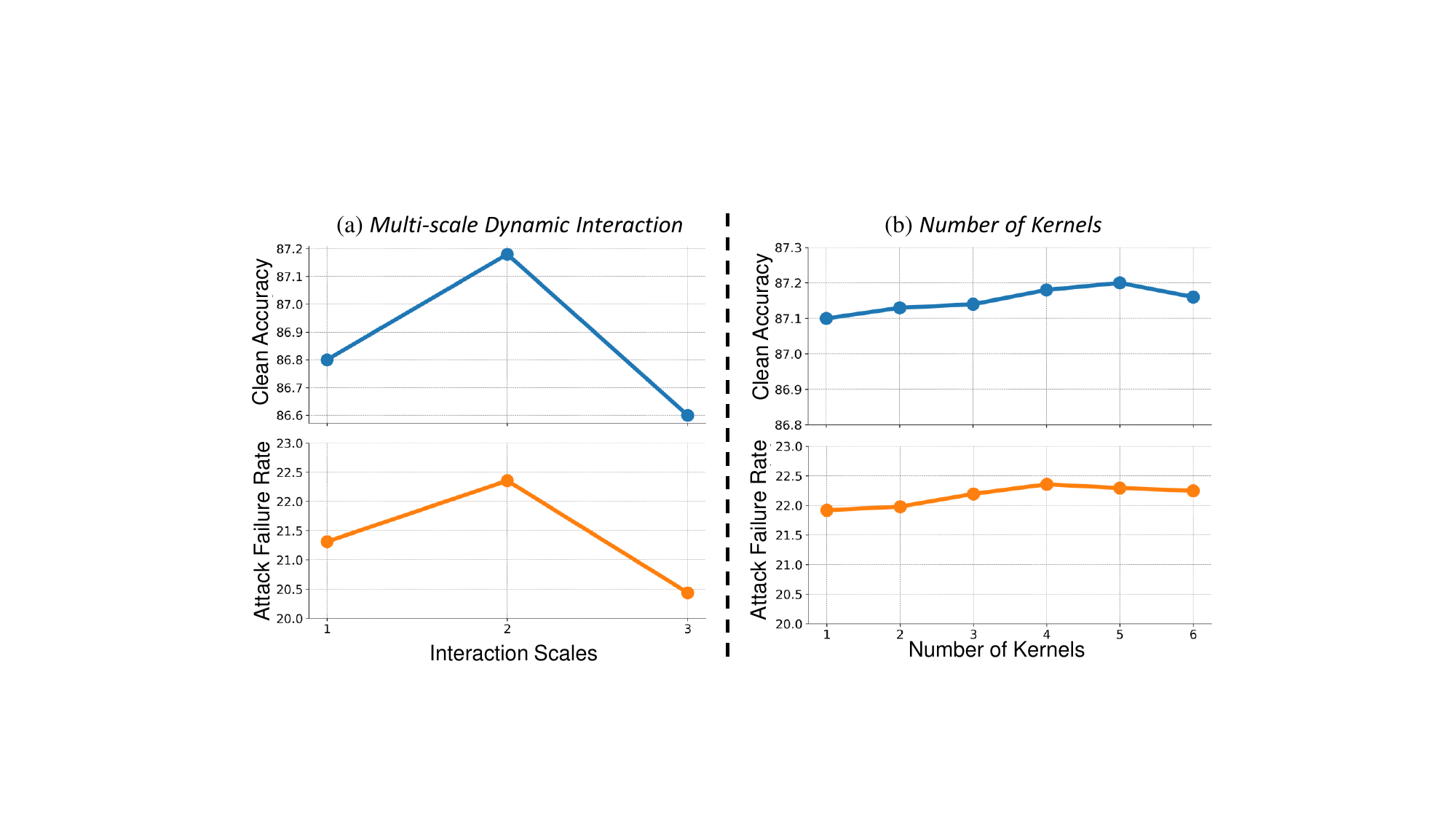}
        \caption{Ablation on the (a) scales and (b) number of kernels used in our multi-scale \textit{\dynperception{}}. \rev{Overly large scales can undermine the performance and so do too many or too few kernels.} We choose $\mathcal{S}=2$, $\mathcal{M}=4$ to balance between the \textcolor{customblue}{clean accuracy} and \textcolor{customorange}{robustness}.}
        \label{fig:abl-msp-kn}
    \end{minipage}
    \vspace{-10pt}
\end{figure}

Our key assumption in Section~\ref{sec:insert-pos} is that \MethodName{} will not change the layer similarity structure due to its small size, and that is why we use \textbf{pre-trained} networks to conduct this analysis. Here, we examine the pair-wise layer similarities in Swin-B \cite{swin} and Swin-B-\MethodName{} in Figure~\ref{fig:layer-sim-swin}. 
$\kappa_{mean}$ is the mean of $\kappa$ from all layer sets. We refer readers to Appendix~\ref{sec:supp-layer-sim-results} for more results. \MethodName{} not only preserves the original layer similarity structure but also contributes to a clearer stage partition, validating our assumption in Section~\ref{sec:insert-pos}. The results might be attributed to the fact that \MethodName{} transmits information between different layer sets and thus layers inside each set do not bother adapting to the layers outside the set. 


\subsection{Ablation studies}
\label{sec:ablation}
We conduct ablation studies on ImageNet100, a 100-class subset of ImageNet1K. Previous studies \cite{multilinear-network,imagenet100-3,imagenet100-1,imagenet100-2} have shown that ImageNet100 serves as a representative subset of ImageNet1K. Hence, we can obtain representative results for the self-evaluation of the model while efficiently using the computational resources. All the ablation experiments are based on a Swin-tiny \cite{swin} model. We insert it after the fourth layer to obtain the best robustness improvement, according to our analysis in Section~\ref{sec:insert-pos} and Appendix~\ref{sec:insert-pos-supp}. First, we ablate on two hyperparameters, the number of convolutional kernels used in the \textit{\dynperception{}} stage ($\mathcal{M}$) and the maximum scale ($\mathcal{S}$) used in the multi-scale \textit{\dynperception{}} stage. The \textcolor{customblue}{Clean Accuracy} and \textcolor{customorange}{Attack Failure Rate} are shown in Figure~\ref{fig:abl-msp-kn}. The attack failure rate is quantified using the same method as in Section~\ref{sec:insert-pos} and Appendix~\ref{sec:insert-pos-supp}. Multi-scale interaction can contribute to the performance while overly large scale can lose local information \rev{and complicate the process of selecting the interaction neighbors. This issue is also observed in a previous work \cite{attention-nca}.} Increasing the number of kernels benefits the performance while too many kernels undermine the robustness. According to the results, we choose $\mathcal{S}=2$, $\mathcal{M}=4$. 
We then perform ablations on several design choices: 
\begin{itemize}[leftmargin=*]
\item \textbf{Recurrent update (Recur).} Ablation on unrolling the recurrence with average time step $T$ in \MethodName{} into $T$ independent \MethodName{} with time step being 1. 
\item \textbf{Stochastic update (StocU).} Ablate the stochastic update during training, leading to globally synchronized update of all tokens \cite{asynchronicity-nca}. 
\item \textbf{Random step (RandS).} Change the recurrence time step from a randomly chosen integer in range $[T_1,T_2]$ to $\left \lceil{ (T_1+T_2)/2}\right \rceil$. It cannot be turned on without recurrence.
\item \textbf{Dynamic interaction (DynIn).} Ablate the \textit{\dynperception{}} so that the interaction results are simply summed together. The number of kernels remains the same. 
\end{itemize}
\rev{As shown in Table~\ref{tab:ablation}, the highest robustness improvement is achieved with all components. Among them, turning off the recurrence leads to the largest drop in the robustness, as recurrence allows the model to explore more cell states than finishing the update in a single step. While it achieves the highest clean accuracy, it uses $\sim$4x more parameters than our method, and the improvement of the clean performance is likely due to more parameters. Without any of the two sources of randomness, stochastic update and random steps, the model cannot adapt to the variability of the inputs and thus exhibits vulnerabilities against adversarial attacks. Finally, turning off our \dynperception{} will cause drops in both clean accuracy and robustness, as tokens cannot decide their unique interaction weights and thus cannot generalize to noisy inputs.  We provide extended ablation studies in Section~\ref{sec:dyn-abl-more} in the Appendix.}

\begin{table*}[]
\caption{Ablation studies on design choices. Each of the \MethodName{} design choices contributes to improved performance and robustness. Baseline is a Swin-tiny \cite{swin} model trained on ImageNet-100. \rev{\textbf{Bold} indicates the best model.}\\}
\label{tab:ablation}
\centering
\resizebox{1.0\linewidth}{!}{
\begin{tabular}{c|cccc|cc|c|c}
\toprule
Exp. Type & Recur & StocU & RandS & DynIn & Params (M) & FLOPS (G) & Accuracy ($\uparrow$)& Attack Failure Rate ($\downarrow$)       \\
\midrule
Baseline & \xmark & \xmark & \xmark & \xmark & 27.59 & 4.5 & 86.56 & 12.29 \\
\midrule
\multirow{4}{*}[0pt]{Ablation} & \xmark & \cmark & \xmark & \cmark & 28.97 & 4.7 & \textbf{87.36} & 19.04 \\
& \cmark & \xmark & \cmark & \cmark & 27.94 & 4.7 & 86.92 & 19.56 \\
& \cmark & \cmark & \xmark & \cmark & 27.94 & 4.7 & 87.12 & 19.34 \\
& \cmark & \cmark & \cmark & \xmark & 27.93 & 4.7 & 86.72 & 21.98 \\
\midrule
Ours & \cmark & \cmark & \cmark & \cmark & 27.94 & 4.7 & 87.18 & \textbf{22.35} \\
\bottomrule
\end{tabular}
}
\vspace{-10pt}
\end{table*}


\subsection{Noise sensitivity examination}
\label{sec:freq-preference}
\begin{figure}
    \centering
    \includegraphics[width=\linewidth, trim={100 190 120 170},clip]{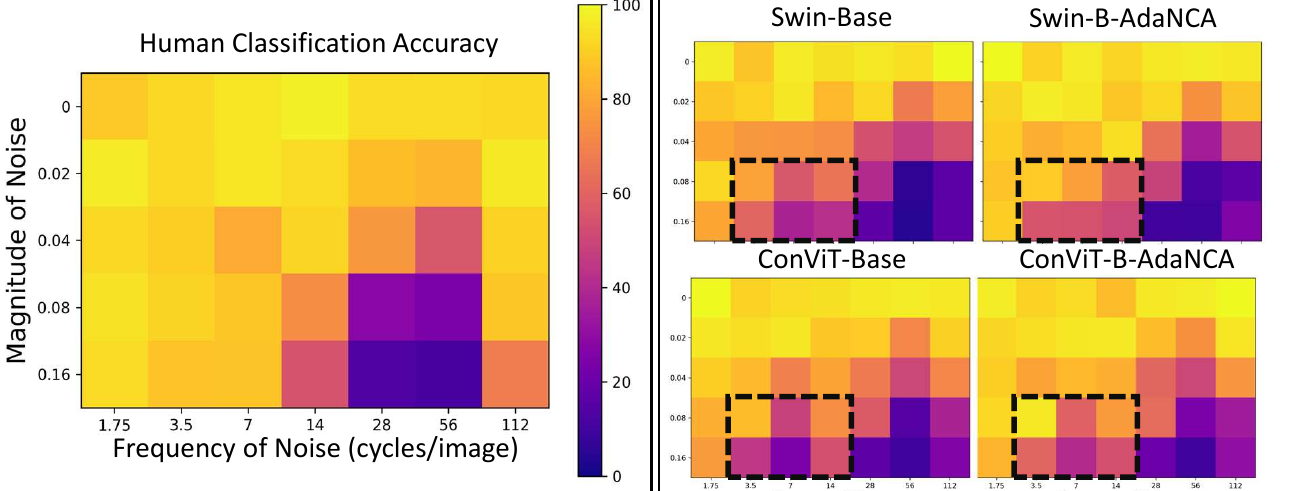}
    \caption{Classification accuracy of humans and ViTs on noisy images. \rev{The images are perturbed by Gaussian noise with different standard deviations (Magnitude of Noise) and are filtered with different spatial-frequency bands (Frequency of Noise).} \MethodName{} improves the accuracy on images with certain types of noise (\textbf{dotted black boxes}), indicating that it makes ViTs less sensitive to them. Quantitative results are presented in Appendix~\ref{sec:supp-freq-result}.}
    \label{fig:freq-bias}
\end{figure}

A drawback of adversarially robust models is their increased sensitivity to noise in specific frequency bands \cite{freq-preference}. 
For instance, while adversarially trained models are robust against adversarial attacks, they can be sensitive to noise from a larger frequency-band range compared to standard models \cite{freq-preference}. Here, we use the method and data from \cite{freq-preference} to examine the noise sensitivity of \MethodName{}-enhanced ViTs. Specifically, we evaluate the classification performance on a set of images that are mixed with noise of varying magnitudes and frequencies. 
\rev{The noise is sampled from Gaussian distribution with zero mean and the standard deviation indicates its magnitude. Then, it is filtered within
various spatial-frequency bands, resulting in different frequencies of noise.}
Higher classification accuracy on a specific noise type indicates that the model is less sensitive to that noise. Figure~\ref{fig:freq-bias} presents the results, including human data sourced from \cite{freq-preference}. Our findings demonstrate that \MethodName{} enhances ViTs by reducing the sensitivity to noise with certain frequency components, equipping ViTs with more human-like noise-resilient abilities. Crucially, \MethodName{} improves the model robustness differently than adversarial training since the \MethodName{}-enhanced ViTs do not exhibit increased sensitivity to the noise. We quantitatively validate the conclusion and refer readers to Appendix~\ref{sec:supp-freq-result} for the details.

\section{Limitation}
\label{sec:limitation}
\FullMethodName{} has certain limitations. First, the \MethodName{}-equipped ViTs cannot adapt to unseen recurrent steps of \MethodName{}, which limits the generalization ability. For example, if the training step range of \MethodName{} is [3,5], it cannot produce meaningful results with the test step being 6.  \MethodName{} introduces a non-negligible computation into the original architecture. Our experiments are conducted on ImageNet1K with the image size of 224 $\times$ 224. Whether \MethodName{} can lead to impressive improvements on larger-scale problems, \textit{e.g.,} ImageNet22K, remains a question. The size of the images can also affect the efficiency of token interaction.

\section{Broader Impact}
\label{sec:broader-impact}
\MethodName{} contributes to more robust ViTs, facilitating their usage in real-world scenarios. We bridge two powerful models, NCA and ViT, on large-scale image classification, potentially encouraging research dedicated to their synergistic combination in more practical settings. Our findings on \MethodName{} improving network redundancy can stimulate more works on architectural robustness in deep learning that involves increasing the redundancy to enhance robustness. \rev{In the context of this paper, we believe that \MethodName{} does not introduce any significant negative implications.}

\section{Conclusion}
We have proposed \MethodName{}, an efficient Neural Cellular Automata (NCA) that, when inserted between their middle layers, improves ViT performances and robustness against adversarial attacks as well as out-of-distribution inputs. We design our model by connecting NCA and ViT, in terms of token interaction modeling, and we propose \textit{\dynperception{}} to improve the computational efficiency of standard NCA. Exploiting the training strategies and design choices in NCA, \textit{i.e.,} stochastic update, random steps, and multi-scale interaction, we further improve the \MethodName{}-enhanced ViTs' clean accuracy and robustness. To decide the placement of \MethodName{}, we propose the Set Cohesion Index that quantifies the network redundancy via layer similarity and conclude that \MethodName{} should be inserted between two layer sets that consist of redundant layers. Our results demonstrate that \MethodName{} consistently improves ViTs performances and robustness. Evidence suggests that the mechanism by which we obtain improvement reduces the sensitivity of ViTs to certain types of noise and makes the noise-resilient ability of ViTs similar to that of humans.

\clearpage
\bibliographystyle{plainnat}
\bibliography{neurips_2024}

\clearpage

\appendix
\section*{Appendix: Table of Contents}
\startcontents[sections]
\printcontents[sections]{}{1}{}
\clearpage
\section{Establish the correlation between \MethodName{} placement and robustness improvement}
\label{sec:insert-pos-supp}
Our motivation for investigating the relationship between \MethodName{} and robustness stems from an empirical observation, namely that making neural networks more monolithic contributes to their robustness \cite{imagenet-c}. One of these monolithic ideas is redundancy in networks. It has been observed that several consecutive layers output similar results \cite{layer-redundancy-robustness}, hence building what we call ``a layer set."
We hypothesize that \MethodName{}, as adaptors inside ViTs, should connect the different sets inside a ViT and transmit information between them. In this way, layers inside a set will no longer bother adapting to other layers outside the set, improving the redundancy and thus robustness. To this end, we propose the layer redundancy quantification $\kappa$ and network redundancy quantification $\mathcal{K}$, as well as the robustness improvement measurement $\gamma$ in Section~\ref{sec:insert-pos}. 

For the robustness improvement, we insert \MethodName{} into all possible positions of 3 ViT models, Swin-tiny \cite{swin}, FAN-small-hybrid \cite{fan}, and RVT-small-plus \cite{rvt}, and train them along with the baseline models on the ImageNet100 dataset \cite{imagenet100-3,imagenet100-1,imagenet100-2}, a representative 100-class subset of ImageNet1K. The total amount of models is 34, including 3 baselines and 31 \MethodName{} models. All \MethodName{}-models have a 1\%-2\% parameter increase and 5\% more FLOPS. The training hyperparameters are given in Table~\ref{tab:train-params-im100}. All models are trained for 300 epochs with the Cosine scheduler. Here, we only consider adversarial robustness for simplification, as it is shown that adversarial robustness is correlated with image corruption robustness within one backbone architecture \cite{rvt}. We use a white-box version of AutoAttack \cite{autoattack} in the \MethodName{} placement analysis in Section~\ref{sec:insert-pos}, which comprises  PGD \cite{pgd}, CW \cite{cw}, APGD-DLR \cite{autoattack}, and APGD-CE \cite{autoattack}. The hyperparameters of the four attacks are in Table~\ref{tab:attack-im100}.

\begin{table}[]
    \centering
    \begin{tabular}{c|ccc}
        \toprule
                      & RVT-S-\MethodName{} & FAN-S-\MethodName{} & Swin-T-\MethodName{} \\
        \midrule
        Learning rate & 2.5e-4                & 2.5e-4                & 2.5e-4                 \\
        Batch size    & 256                & 256                & 256                 \\
        Model EMA decay     & 0.99992             & 0.9999             & /                 \\
        Stochastic depth     & 0.1                 & 0.25                & 0.1           \\
        Random erase prob     & 0.25         & 0.3                 & 0.25                \\
        Gradient clip & None                & None                & 5.0                  \\
        MixUp & 0.8                & 0.8                 & 0.8                  \\
        Label smoothing & 0.1                & 0.1                 & 0.1                  \\
        Min learning rate & 1e-5                & 1e-6                & 1e-5                   \\
        Weight decay & 0.05                & 0.05                 & 0.05                  \\
        \bottomrule
    \end{tabular}
    \caption{Hyperparameters used in \MethodName{}-equipped ViT training for analysis in Section~\ref{sec:insert-pos} on ImageNet100.}
    \label{tab:train-params-im100}
\end{table}

\begin{table}[]
    \centering
    \begin{tabular}{c|cccc}
        \toprule
                      & PGD \cite{pgd} & CW \cite{cw} & APGD-DLR \cite{autoattack} & APGD-CE \cite{autoattack}  \\
        \midrule
        Max magnitude & 0.5                & /               & 0.5                 & 0.5 \\
        Steps         & 5                  & 20                & 5                   & 5   \\
        Step size         & 0.25                  & /                & /                   & /   \\
        $\kappa$         & /                   & 0.0                 & /                   & /   \\
        $c$         & /                   & 0.25                & /                   & /   \\
        $\rho$         & /                   & /                 & 0.75                   & 0.75   \\
        EoT         & /                   & /                 & 1                   & 1   \\
        \bottomrule
    \end{tabular}
    \caption{Hyperparameters in different adversarial attacks used in the analysis in Section~\ref{sec:insert-pos} on ImageNet100 as well as our ablation studies.}
    \label{tab:attack-im100}
\end{table}

\begin{figure}
    \centering
    \includegraphics[width=\linewidth,trim={105 60 110 80},clip]{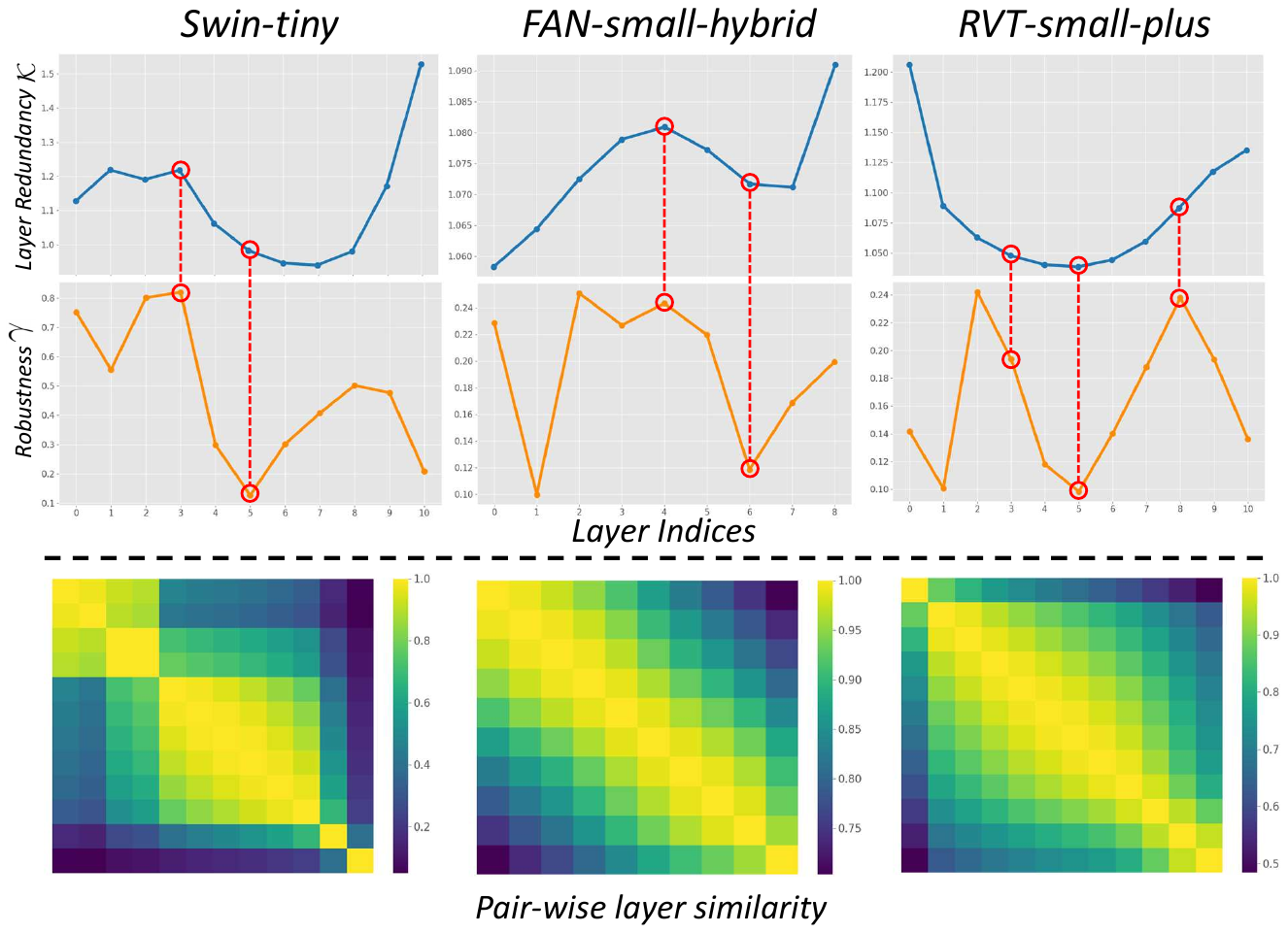}
    \caption{Top: Visualization of layer redundancy ($\mathcal{K}(i)=\kappa(1,i+1)+\kappa(i+2,L)$) and corresponding robustness improvement $\gamma$. Bottom: Visualization of pair-wise layer similarity of the 3 ViTs.}
    \label{fig:kappa-layer-auto-supp}
\end{figure}

\begin{figure}
    \centering
    \includegraphics[width=\linewidth,trim={20 105 70 90},clip]{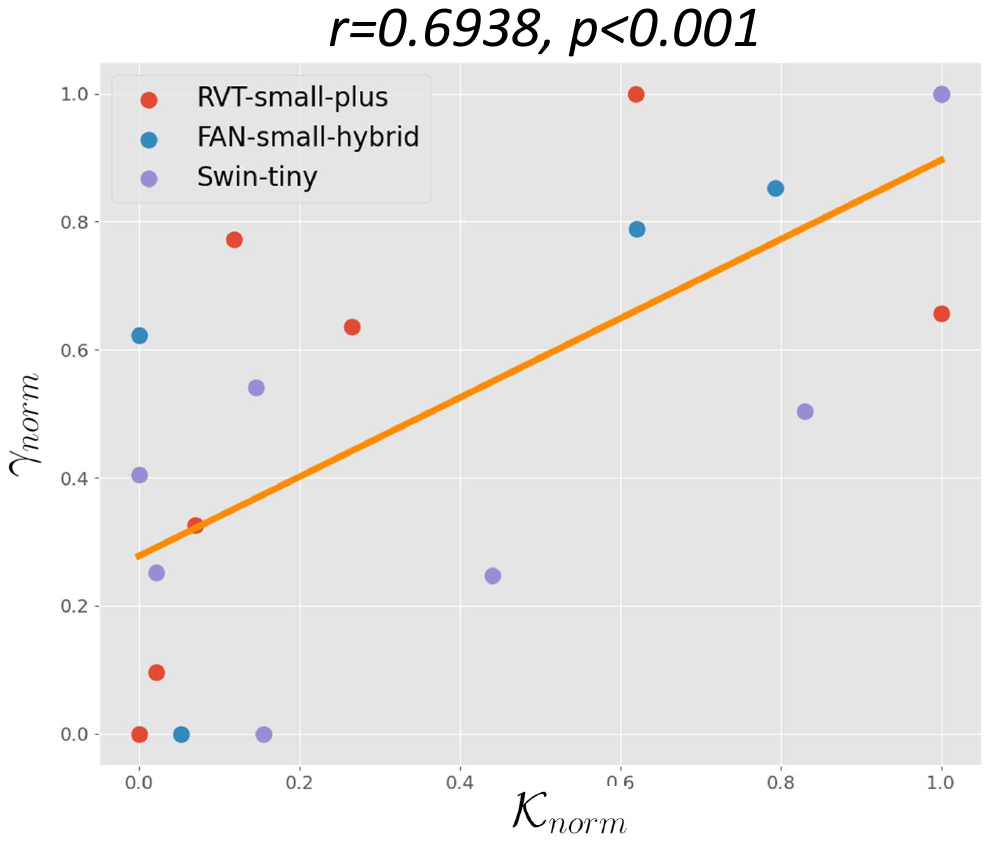}
    \caption{The relationship between the layer redundancy $\mathcal{K}$ and robustness improvement $\gamma$ by inserting \MethodName{} in the corresponding position. $\gamma_{norm}$ and $\mathcal{K}_{norm}$ are obtained by normalizing $\gamma$ and $\mathcal{K}$ within the results of each model. Note that the coordinate (1.0,1.0) has two overlapping points (Swin-tiny and FAN-small-hybrid). Quantitatively, $\gamma_{norm}$ is significantly correlated with $\mathcal{K}_{norm}$ ($r=0.6938,p<0.001$). The linear fit is shown in orange.}
    \label{fig:kappa-corr-supp-quant}
\end{figure}

We show the qualitative results of our experiments on the Set Cohesion Index and robustness improvement in Figure~\ref{fig:kappa-layer-auto-supp} and the quantitative results in Figure~\ref{fig:kappa-corr-supp-quant}. Qualitatively, $\gamma$ generally follows the trend of $\mathcal{K}$ except for the top 3 layers and the last layer. Specifically, $\gamma$ for the top three layers consistently exhibits a pattern where it decreases in the second layer and increases in the third layer, regardless of the trend in $\mathcal{K}$, which always increases.
We hypothesize that the proximity of the top 3 layers to the input image causes \MethodName{} to similarly impact the model's robustness by adapting the input to the subsequent layers. Furthermore, the position immediately before the final layer likely serves as a transitional stage for output, adapting to different output strategies (e.g., FAN has an additional class attention head while Swin and RVT use average pooling on all tokens to generate features for classification). The distinct behavior of the last layer compared to other layers has also been noted in previous research \cite{no-deep-layer}.
Hence, we exclude the models (\MethodName{} applied in the top three layers and before the final layer) from our analysis. The resulting number of models is 19. To conduct a cross-model quantitative comparison, we normalize all $\gamma$ as well as all $\mathcal{K}$ in a single type of model. Each of the 3 models thus has a set of $\gamma_{norm} \in [0,1]$ as well as $\mathcal{K}_{norm} \in [0,1]$. We plot all sets of $\gamma_{norm}$ and $\mathcal{K}_{norm}$ from 3 models in Figure~\ref{fig:kappa-corr-supp-quant}.  Note that the coordinate (1.0,1.0) has two overlapping points (Swin-tiny and FAN-small-hybrid), hence only 18 points are visible in the figure. $\gamma_{norm}$ is significantly correlated with $\mathcal{K}_{norm}$ ($r=0.6938,p=9.8e-4$), as we report in Section~\ref{sec:insert-pos}.

\subsection{Dynamic programming for \MethodName{} placement}
\label{sec:dynamic-programming}
Given $\mathcal{U}$ \MethodName{}s, the ViT is expected to be partitioned into $\mathcal{U}+1$ sets. The dynamic programming involves filling an array $\mathcal{D}$, where $\mathcal{D}[i][u]$ represent the maximum value of $\mathcal{K}$ achievable by partitioning the first $i$ layers into $u$ sets. The $\mathcal{D}$ can be obtained via Equation~\ref{eq:dp-stageness}.
\begin{equation}
    \mathcal{D}[i][u] = \mathop{max}_{j<i}(\mathcal{D}[j][u-1]+\kappa(j,i-1))
    \label{eq:dp-stageness}
\end{equation}
The boundary conditions are: 1) $\mathcal{D}[i][1]=\kappa(1,i)$, which is partitioning the first $i$ layers into 1 set; 2) $\mathcal{D}[0][u]=0$. By recording the partition position, we can find $\mathcal{U}$ partition points for inserting the \MethodName{}. The pseudo-code is presented in Algorithm~\ref{alg:kappa}.
\begin{algorithm}[H]
    \caption{find\_optimal\_partition(Similarity Matrix \( S \), Number of Stages \( \mathcal{U} \))}
    \begin{algorithmic}[1]
        \STATE \( n \gets \text{length of } S \)
        \STATE Initialize \( \mathbf{dp} \) with dimensions \((n+1, \mathcal{U}+1)\), filled with \(-\infty\)
        \STATE Initialize \( \mathbf{partition} \) with dimensions \((n+1, \mathcal{U}+1)\), filled with zeros
        \STATE \( \mathbf{dp}[0][0] \gets 0 \) \COMMENT{Base case}
        
        \FOR{each \( i \) from 1 to \( n \)}
            \FOR{each \( u \) from 1 to $\min$(i, $\mathcal{U}$)}
                \FOR{each \( j \) from 0 to \( i-1 \)}
                    \STATE \( \text{current\_value} \gets \mathbf{dp}[j][u-1] + \kappa(j,i-1) \)
                    \IF{\( \text{current\_value} > \mathbf{dp}[i][u] \)}
                        \STATE \( \mathbf{dp}[i][u] \gets \text{current\_value} \)
                        \STATE \( \mathbf{partition}[i][u] \gets j \)
                    \ENDIF
                \ENDFOR
            \ENDFOR
        \ENDFOR
        
        \STATE Initialize \( \text{stages} \) as an empty list
        \STATE \( \text{current\_layer} \gets n \)
        \STATE \( \text{current\_stage} \gets \mathcal{U} \)
        \WHILE{\( \text{current\_stage} > 0 \)}
            \STATE \( \text{start\_layer} \gets \mathbf{partition}[\text{current\_layer}][\text{current\_stage}] + 1 \)
            \STATE Append \((\text{start\_layer}, \text{current\_layer})\) to \( \text{stages} \)
            \STATE \( \text{current\_layer} \gets \mathbf{partition}[\text{current\_layer}][\text{current\_stage}] \)
            \STATE \( \text{current\_stage} \gets \text{current\_stage} - 1 \)
        \ENDWHILE
        \STATE stages = stages[::-1]
        \RETURN \text{stages}
    \end{algorithmic}
    \label{alg:kappa}
\end{algorithm}
Note that here the layer index starts from 0, and so does the input requirement of $\kappa$.

\section{Notes on statistical significance}
\label{sec:errorbar}
We do not report the error bar for training models with different seeds on ImageNet1K, as it would be very expensive (Table~\ref{tab:train-time}). However, we follow the seed used in the released code for each model. We test the adversarial attack using 5 seeds and find the difference between different seeds negligible (standard deviation: PGD 0.08, CW 0.01, APGD-DLR 0.01, APGD-CE 0.02). As all our results on adversarial attacks have differences larger than $0.1$, they are statistically significant. 

For ImageNet100 experiments in the ablation study, we train the model 3 times and find the clean accuracy changes within a small range (standard deviation $0.08$), and the robustness test results barely change (standard deviation $0.02$). We also test the robustness improvement within one model using 5 seeds and the standard deviation is $0.008$. 

For the \MethodName{} placement analysis, our correlation result is statistically significant ($p<0.001$). 

\section{ImageNet1K experiments}
\label{sec:im1k-supp}
\subsection{\MethodName{} settings for each model}
\label{sec:adanca-placement-detail}
Based on Algorithm~\ref{alg:kappa}, we insert \MethodName{} into the four chosen baseline models. The specific settings for each \MethodName{} are shown in Table~\ref{tab:adanca-setting}. Note that insert position 0 means before the network, since FAN-B model uses a complex patch embedding layer, we can treat it as a unique layer that encodes semantic information rather than the simple convolution patch embedding used in other models. In practice, we find inserting \MethodName{} after it can contribute to the model performance. Moreover, we add the drop path operation after each \MethodName{}, and the drop path rate follows the one of the layer that \MethodName{} follows. Hence, the regularization is pretty strong even without the stochastic update. However, as shown in our ablation studies in Section~\ref{sec:ablation}, the stochastic update indeed contributes to the model performance. Our FLOPS computation for \MethodName{} models are based on Test steps.
\begin{table}[]
    \caption{\MethodName{} settings for each ViT.}
    \label{tab:adanca-setting}
    \centering
    \resizebox{\linewidth}{!}{
    \begin{tabular}{c|ccccc}
    \toprule
                        & \multirow{2}{*}{\# Layers} & \MethodName{} positions & \multirow{2}{*}{Training steps} & \multirow{2}{*}{Test steps} & \multirow{2}{*}{Stochastic update}  \\
                        & & (after the index) & & & \\
    \midrule
    RVT-B-\MethodName{} & 12 & 4,8 & [2,4],[3,5] & 3,3 & 0.1  \\
    FAN-B-\MethodName{} & 16 & 0,6,9 & [2,2],[3,5],[2,4] & 2,4,3 & 0.1  \\
    Swin-B-\MethodName{} & 24 & 2,8,22 & [2,2],[3,5],[2,4] & 2,4,3 & 0.1  \\
    ConViT-B-\MethodName{} & 12 & 5,9 & [3,5],[2,4] & 3,3 & 0.1  \\
    \bottomrule
    \end{tabular}
    }
\end{table}

\subsubsection{Discussion on the choice of \MethodName{} steps}
\rev{Our choice of \MethodName{} steps differs from all previous NCA models, which typically iterate for hundreds of steps. We make this compromise as the computational costs increase linearly as the step grows. We aim to minimize the increase in the number of parameters and FLOPS for scalability and we do not want the source of improvement to merely stem from the increase in the size and computation of the models. We show that more NCA steps will indeed contribute to the model's performance in Table~\ref{tab:increase-step}, while it introduces extra FLOPS. The model is the same as in Section~\ref{sec:ablation} which is trained using a range of steps being $[3,5]$. We want to underscore that it is the architecture and evolution scheme that defines an NCA as introduced in Section~\ref{sec:preliminary}, instead of the number of its recurrent steps. Admittedly, fewer steps will lead to a coarser path to the target state, and can potentially undermine the model's performance. Notably, with our scheme, \MethodName{} has achieved generally strong robustness improvements compared to the baselines. It indicates that our choice of the recurrent steps of \MethodName{} is a good balance between computational costs and performance. }

\begin{table}[]
\centering
\caption{Increasing the number of recurrent steps of \MethodName{} during testing can contribute to the model performance.}
\label{tab:increase-step}
\begin{tabular}{c|cccc}
\toprule
       & \# Params (M) & FLOPS (G) & Clean Acc. ($\uparrow$) & Attack Failure Rate ($\uparrow$) \\
       \midrule
Step=4 & 27.94         & 4.7       & 87.18                   & 22.35                            \\
Step=5 & 27.94         & 4.8       & 87.22                   & 23.46   \\
\bottomrule
\end{tabular}
\end{table}

\subsection{Discussion on the model choices}
In our experiments, we choose RVT \cite{rvt}, FAN \cite{fan}, Swin \cite{swin}, and ConViT \cite{d2021convit}. Our choice not only includes different types of ViTs (Regular, Hybrid, Hierarchical) and robust as well as non-robust architectures (RVT, FAN and Swin, ConViT), but also covers other aspects of ViT characteristics. First, RVT and Swin use average pooling on the final token maps to obtain the 1D feature for classification, while FAN and ConViT adopt a separate class token. It has been proved that those two strategies of classification can have a significant influence on the model robustness \cite{rvt}. Despite this, \MethodName{} are effective in both ways for classification. 
Moreover, all the architectures already contain certain kinds of local structures. RVT contains depth-wise convolution within MLP and a convolutional patch embedding layer. FAN has a ConvNeXt \cite{convnext} head for patch embedding. Swin has shifted window attention. ConViT has convolution-attention coupled gated positional self-attention layers. Local structures have been proven effective in improving model robustness \cite{rvt}. Therefore, we partially eliminate the possibility that \MethodName{} is effective simply because of introducing local information in ViTs without any local inductive bias, such as the original ViT \cite{vit} or DeiT \cite{deit}. As shown in our ablation studies in Section~\ref{sec:ablation}, \MethodName{} not only contributes to more parameter-efficient models but also improves the model robustness compared to not using the training strategies or design from NCA. 
Furthermore, in our choices, Swin does not conduct weight exponential moving average (EMA), while the other 3 models perform. It indicates that \MethodName{} can be effective with or without model EMA.

\subsection{Discussion on the hyperparameters of MLP inside \MethodName{}}
In \MethodName{}, we set the hidden dimensionality of the MLP to the same one as the input dimension. It is different from the common design choice of MLP in ViT that uses a 4-time larger hidden layer than the input layer. Moreover, it also deviates from the design of the previous NCA that uses more than 8 times larger hidden layer size than the input one. Instead, we set the hidden dimensionality to be the same as the input one. The reason is that \MethodName{} introduces additional parameters and computation in ViTs and we want the extra computational overhead to be as low as possible. The lowest dimensionality that will not cause inevitable information loss is the same one as the input dimension. Therefore, we design the MLP in \MethodName{} as a non-compression-non-expand structure. 

\subsection{The effect of \MethodName{} placement on robustness on ImageNet1K}
Our algorithm for deciding the placement of \MethodName{} is based on the correlation between the robustness improvement and the Set Cohesion Index in Section~\ref{sec:insert-pos}. All the experiments are conducted on ImageNet100 to efficiently use the computational resources. Here, we compare two schemes for deciding the placement of \MethodName{} on ImageNet1K on Swin-B \cite{swin} and FAN-B \cite{fan} to demonstrate our result validity. The first scheme is \textbf{No-Prior}, which does not involve the knowledge of the layer similarity and performs the most reasonable placement choice. For Swin-B, the choice for placing 3 \MethodName{} is to insert them between each stage pair defined by the transition between different embedding dimensionalities, namely after layer 2, layer 4, and layer 22. Coincidentally, this choice aligns with what we obtain from the dynamic programming algorithm, which is to place \MethodName{} after layer 2, layer \textbf{8}, and layer 22. For FAN-B, although it is a hybrid ViT, we only consider inserting \MethodName{} into its main network where all layers share the same structure. In this case, the \textbf{No-Prior} chooses to uniformly insert \MethodName{} between the 16 layers, that is after layers 5 and 10. Our algorithm decides the placement should instead be after layers 6 and 9. Table~\ref{tab:supp-swin-placement} shows the results of two Swin-B-\MethodName{} models. Inserting \MethodName{} with the \textbf{No-Prior} scheme can still contribute to the model performance and robustness, but not as effective as using our proposed method. 

\begin{table*}[]
\caption{Comparison between different schemes of \MethodName{} placement. \textbf{No-Prior}: insert \MethodName{} without knowledge of layer similarity structure but making the most reasonable choice, that is to insert them between the architectural stage transition where the embedding dimensionality changes for Hierarchical ViT like Swin \cite{swin}, or insert them uniformly between layers that have the same structure like the main network in FAN \cite{fan}. \MethodName{} consistently improves the model performance and robustness compared to the baseline models, while the improvement can be maximized by using our \MethodName{} placement scheme based on Algorithm~\ref{alg:kappa}.\\}
\label{tab:supp-swin-placement}
\resizebox{\linewidth}{!}{
\begin{tabular}{c|c|cc|c|cccccccc}
\toprule
\multirow{2}{*}{\textbf{Model}} &  \multirow{2}{*}{\textbf{Insert Scheme}} &  Params & FLOPS  & \textbf{ImageNet} & \multicolumn{8}{c}{\textbf{Robustness Benchmarks}}\\
                             &  & (M) & (G) & Top-1 & PGD & CW & APGD-DLR & APGD-CE & IM-C ($\downarrow$) & IM-A & IM-R & IM-SK \\
\midrule
Swin-B \cite{swin}         & /   & 87.8 & 15.4 & 83.4 & 21.3 & 13.4 & 15.6 & 23.1 & 54.2 & 35.8 & 46.6 & 32.4 \\
\textit{Swin-B-\MethodName{}} & No-Prior & 90.7 & 16.3 & 83.6 & 23.1 & 19.4 & 24.8 & 24.5 & 52.0 & \textbf{36.0} & 47.7 & 34.8 \\
\rowcolor{gray!20} \textit{Swin-B-\MethodName{}} & \textit{Ours} & 90.7 & 16.3 & \textbf{83.7} & \textbf{24.1} & \textbf{20.5} & \textbf{25.1} & \textbf{24.8} & \textbf{51.5} & \textbf{36.0} & \textbf{48.2} & \textbf{35.5} \\
\midrule
FAN-B \cite{fan}                  & /             & 50.4 & 11.7 & 83.9 & 15.0 & 7.6 & 10.4 & 13.1 & 46.1 & 39.6 & 52.7 & 40.8 \\
\textit{FAN-B-\MethodName{}}  &  No-Prior   & 51.7 & 12.4 & \textbf{84.1} & 17.3 & 10.2 & 13.5 & 16.3 & \textbf{44.7} & 42.8 & \textbf{53.5} & \textbf{41.0} \\
\rowcolor{gray!20} \textit{FAN-B-\MethodName{}}     & \textit{Ours}              & 51.7 & 12.4 & \textbf{84.1} & \textbf{20.3} & \textbf{10.6} & \textbf{14.1} & \textbf{19.1} & \textbf{44.7} & \textbf{42.9} & 53.4 & \textbf{41.0} \\
\bottomrule
\end{tabular}
}
\end{table*}

\subsection{Details of architecture change in Swin-Base$^\star$ and ConViT-Base$^\star$}
For Swin-Base$^\star$, we add two extra layers in stage 3 with embedding dimensionality being 512, since in the original design most of the computational budgets are dedicated to stage 3, and two layers form a complete Swin operation. For ConViT-Base$^\star$, we add an extra Gated Positional Self-Attention (GPSA) layer. 

\subsection{Training details}
\label{sec:train-detail}
We use different training schemes for the four selected baseline models, closely following the parameter settings for each of them (link to reference configuration files: \textcolor{red}{\href{https://github.com/vtddggg/Robust-Vision-Transformer?tab=readme-ov-file\#rvt-b}{RVT-B}},\textcolor{red}{\href{https://github.com/NVlabs/FAN/blob/master/scripts/fan_vit/fan_net_base.sh}{FAN-B}},\textcolor{red}{\href{https://github.com/microsoft/Swin-Transformer/blob/main/configs/swin/swin_base_patch4_window7_224.yaml}{Swin-B}},\textcolor{red}{\href{https://github.com/facebookresearch/convit?tab=readme-ov-file\#training}{ConViT-B}},). Most training settings are in line with DeiT \cite{deit} since all models used build the training upon the scheme of training a DeiT but with minor changes. Some specific training settings are listed in Table~\ref{tab:train-params}.

\begin{table}[]
    \caption{Hyperparameters used in \MethodName{}-equipped ViT training.}
    \label{tab:train-params}
    \centering
    \resizebox{\linewidth}{!}{
    \begin{tabular}{c|cccc}
        \toprule
                      & RVT-B-\MethodName{} & FAN-B-\MethodName{} & Swin-B-\MethodName{} & ConViT-B-\MethodName{}  \\
        \midrule
        Learning rate & 1e-3                & 2e-3                & 1e-3                 & 1e-3 \\
        Batch size    & 1024                & 2048                & 1024                 & 1024 \\
        Model EMA decay     & 0.99996             & 0.99992             & /                  & 0.99996 \\
        Stochastic depth     & 0.1                 & 0.35                & 0.5           & 0.1 \\
        Random erase prob     & 0.25         & 0.3                 & 0.25                & 0.25 \\
        Gradient clip & None                & None                & 5.0                  & None \\
        MixUp & 0.8                & 0.8                 & 0.8                  & 0.8  \\
        Label smoothing & 0.1                & 0.1                 & 0.1                  & 0.1  \\
        Min learning rate & 1e-5                & 1e-6                & 1e-5                  & 1e-5  \\
        Weight decay & 0.05                & 0.05                 & 0.05                  & 0.05  \\
        \bottomrule
    \end{tabular}
    }
\end{table}
All models are trained for 300 epochs. Our ablation studies on the size of the model (Swin-B-abl, ConViT-B-abl) also follow the same training settings. The detailed time consumption for training each model using 4 Nvidia-A100 80G GPUs is in Table~\ref{tab:train-time}. The excessive time of RVT is because of the patch-wise augmentation \cite{rvt}. 

\begin{table}[]
    \caption{Training time of each \MethodName{}-equipped ViT.}
    \label{tab:train-time}
    \centering
    \resizebox{\linewidth}{!}{
    \begin{tabular}{c|cccc}
        \toprule
                      & RVT-B-\MethodName{} & FAN-B-\MethodName{} & Swin-B-\MethodName{} & ConViT-B-\MethodName{}  \\
        \midrule
        Training time (hours) & 400                & 180               & 120                 & 100 \\
        \bottomrule
    \end{tabular}
    }
\end{table}

\subsection{Adversarial attacks}
\label{sec:supp-adv-detail}
We test all \MethodName{}-equipped ViTs using four adversarial attacks based on the code \cite{attack-github}. The detailed setting of each attack is given in Table~\ref{tab:attack-im1k}. We try setting the Expectation Over Transformation (EOT) to 3 and the results do not change significantly (For Swin-B, APGD-DLR, EOT=1: 25.124, EOT=3: 25.121). Hence, we fix the EOT to 1.
\begin{table}[]
    \caption{Hyperparameters in different adversarial attacks.\\}
    \label{tab:attack-im1k}
    \centering
    \begin{tabular}{c|cccc}
        \toprule
                      & PGD \cite{pgd} & CW \cite{cw} & APGD-DLR \cite{autoattack} & APGD-CE \cite{autoattack}  \\
        \midrule
        Max magnitude & 1.0                & /               & 1.0                 & 1.0 \\
        Steps         & 5                  & 20                & 5                   & 5   \\
        Step size         & 0.5                  & /                & /                   & /   \\
        $\kappa$         & /                   & 0.0                 & /                   & /   \\
        $c$         & /                   & 0.5                & /                   & /   \\
        $\rho$         & /                   & /                 & 0.75                   & 0.75   \\
        EOT         & /                   & /                 & 1                   & 1   \\
        \bottomrule
    \end{tabular}
\end{table}
\subsubsection{Choice of adversarial attacks}
We choose PGD \cite{pgd} since it is the most popular method for examining model adversarial robustness after architectural changes \cite{attention-sparse-recurrent,rvt}. However, it is relatively easy to overcome the PGD attack using obfuscated gradients through methods such as random inference, noisy architecture, or non-differentiable components \cite{obfuscated-grad}. Our \MethodName{} does not fulfill the requirement for producing obfuscated gradients since it is fully differentiable and we turn off all randomness during test. However, we still want to examine whether recurrence would result in corrupted gradient information due to the drawback of cross entropy loss \cite{autoattack}. Moreover, the step size in PGD can largely affect the result. Hence, we choose Auto-PGD family \cite{autoattack} to automatically decide the step size and incorporate the new Difference of Logits Ratio (DLR) loss, resulting in APGD-CE and APGD-DLR, respectively. We also want to include an optimization-based adversarial attack and thus select the CW \cite{cw} attack. 

\subsubsection{Black-box attack}
We also consider the black-box attack, specifically Square \cite{square}. However, due to the extreme computational cost of performing Square on ImageNet1K, we instead conduct Square attack to three models, Swin-tiny (Swin-T) \cite{swin}, FAN-small-hybrid (FAN-S) \cite{fan}, and RVT-small-plus (RVT-S), on ImageNet100. The three models are used in our \MethodName{} analysis in Section~\ref{sec:insert-pos}. The maximum magnitude $\epsilon=6/255$ and the number of queries is 1000. Results are shown in Table~\ref{tab:square-attack}. The \MethodName{} placements are in line with the highest robustness improvement placements demonstrated in Figure~\ref{fig:kappa-layer-auto-supp}.

\begin{table}[]
    \caption{Results of ViTs and \MethodName{} models under Square \cite{square} attack on ImageNet100.\\}
    \label{tab:square-attack}
    \centering
    \begin{tabular}{c|cccc}
        \toprule
                     & Placement (after) & Clean Acc. & Square  & Attack Failure Rate \\
        \midrule
        Swin-T \cite{swin} & / & 86.56                & 16.18               & 18.69       \\
        \rowcolor{gray!20} Swin-T-\MethodName{} & 4 & \textbf{87.18}                & \textbf{31.74}               & \textbf{36.41}       \\
        \midrule
        FAN-S \cite{fan} & / & 87.30                & 29.44               & 33.72      \\
        \rowcolor{gray!20} FAN-S-\MethodName{} & 5  & \textbf{88.22}                & \textbf{42.24}               & \textbf{47.88}       \\
        \midrule
        RVT-S \cite{rvt} & / & 87.18                & 33.14               & 38.01       \\
        \rowcolor{gray!20} RVT-S-\MethodName{} & 9  & \textbf{87.50}                & \textbf{39.12}               & \textbf{44.71}       \\
        \bottomrule
    \end{tabular}
\end{table}

\subsection{The effect of the range of the steps in random step training}
\rev{Our random stap training strategy contributes to the model robustness as shown in Table~\ref{tab:ablation}. In all of our experiments, we adopt a range of 2 in the random step setting. Here, we examine the effect of increasing the range. We use the same setting as described in Section~\ref{sec:ablation}. Results are given in Table~\ref{tab:range-test}. We can observe that more choices of the recurrent steps worsen the model performance. This might stem from too much noise introduced into the training process which leads to underfitting. Therefore, we only adopt a range of 2 in our experiments.} 
\begin{table}[]
    \caption{Results of Swin-Tiny and the \MethodName{}-enhanced models with different ranges of random steps on ImageNet100.\\}
    \label{tab:range-test}
    \centering
    \begin{tabular}{c|cccc}
        \toprule
                     & Range & Clean Acc. & AutoAttack  & Attack Failure Rate \\
        \midrule
        Swin-T \cite{swin} & / & 86.56                & 10.64               & 12.29       \\
        \midrule
        \multirow{3}{*}[0pt]{Swin-T-AdaNCA} & [3,5] & \textbf{87.18}                & \textbf{19.52}              & \textbf{22.35}      \\
        & [2,6] & 87.02                & 19.02               & 21.85       \\
        & [1,7] & 86.84                & 16.20               & 18.65       \\
        \bottomrule
    \end{tabular}
\end{table}

\subsection{Reason for using ImageNet22K model for RVT}
\label{sec:supp-notes-on-result}
We use the ImageNet22K-pretrained model for RVT-Base-plus in our main results. This is because we do not find an official release of the ImageNet1K-trained weights. Moreover, our trainable parameter count on the model in the released code differs from the one reported in the original paper (ours: 88.5M, in-paper: 91.8M) \cite{rvt, rvt-github}. Hence, self-training might deviate from the results in the official paper. We tried to contact the author for an ImageNet1K version of the model but did not succeed. We test the result of the released model (ImageNet22K version) on PGD attack and get an accuracy of 30.47. In the paper, the authors report this number to be 29.9 on the ImageNet1K model. Hence, we assume that the results of adversarial attacks can be used as a representative, though might be slightly optimistic. Importantly, \textbf{this does not falsify our claim that TAPADL modification undermines the adversarial robustness of RVT}, since TAPADL-RVT already falls short in PGD attack when compared to the ImageNet1K version of RVT-Base-plus. However, for the O.O.D test, ImageNet22K model can differ a lot from the ImageNet1K models, as most data would be counted as in-distribution w.r.t ImageNet22K. Therefore, we use the official values reported in the original paper \cite{rvt} on those benchmarks. Moreover, for the comparison other than adversarial attacks, we use TAPADL-RVT \cite{robustfy-attention-tapadl} as a proxy of the RVT model.

\subsection{Visualization of attention maps}
\begin{figure}
    \centering
    \includegraphics[width=\linewidth,trim={140 90 150 80},clip]{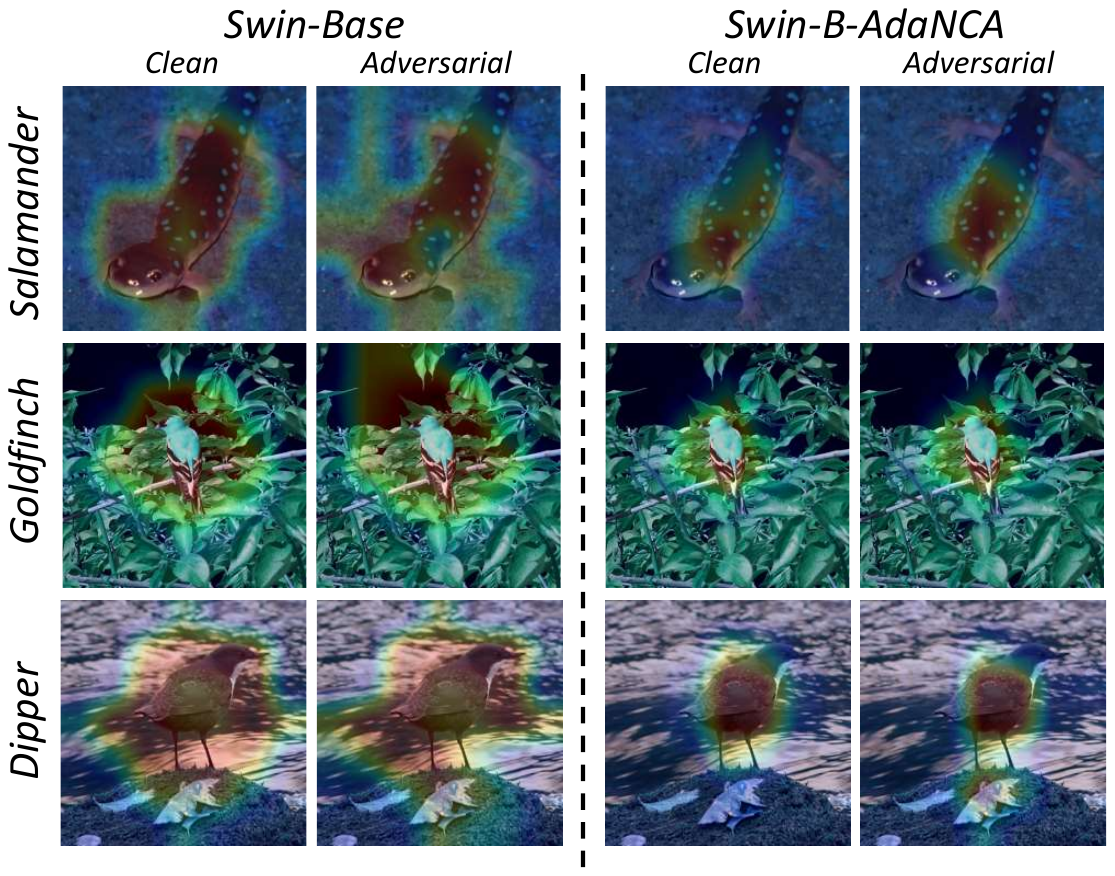}
    \caption{Visualization of attention maps of Swin-Base \cite{swin} and Swin-B-\MethodName{} using GradCAM++ \cite{gradcam-plus-plus} on clean images and images with adversarial noise. \MethodName{} helps ViTs focus more on the object when facing noise.}
    \label{fig:atten-vis}
\end{figure}

We aim to qualitatively show that \MethodName{} helps ViTs perform correct classification when facing noise. Here, we use GradCAM++ \cite{gradcam-plus-plus} to visualize the attention map of Swin-Base \cite{swin} and Swin-B-\MethodName{} on the clean images as well as images containing adversarial noise. We use APGD-DLR to generate the adversarial images. Results are shown in Figure~\ref{fig:atten-vis}. We can observe that \MethodName{}-enhanced Swin model focuses more on the objects, while the baseline model attends to areas unrelated to the object on the images when facing adversarial noise.


\subsection{Integrating AdaNCA into pre-trained ViT models}
\rev{In our experiments, we train all models from scratch. Implementing \MethodName{} as a plug-and-play module for pre-trained ViT models would certainly improve the training speed. To explore this, we experiment by inserting \MethodName{} into a pre-trained Swin-base model on ImageNet1K by 1) freezing all ViT layers; 2) training only the boundary layers, where boundary layers are the layers before and after the insert position of \MethodName{}; 3) Finetuning all layers. Results are given in Table~\ref{tab:finetune-abl}. None of the schemes perform as well as training from scratch. The results indicate that the current NCA is able to adapt to such a scheme. However, it may struggle to effectively transmit information between two pre-trained ViT layers, as these layers have already established strong connections. In contrast, training the model from scratch allows NCA and ViT to synergistically adapt to feature variability, resulting in better overall performance. However, it is worth exploring in the future since the fine-tuning scheme can contribute to the performance.}

\begin{table}[]
\centering
\caption{Integrating \MethodName{} with a pre-trained ViT on ImageNet1K. All integration schemes perform worse than training Swin-B-AdaNCA from scratch.}
\label{tab:finetune-abl}
\begin{tabular}{c|ccccc}
                  & Clean Acc. ($\uparrow$) & IM-A ($\uparrow$) & APGD-DLR ($\uparrow$) & IM-R ($\uparrow$) & IM-SK ($\uparrow$) \\
                  \toprule
Swin-B            & 83.4       & 35.8 & 15.6     & 46.6 & 32.4  \\
\rowcolor{gray!20}  \textit{Swin-B-AdaNCA}   & \textbf{83.7}       & \textbf{36.0} & \textbf{25.1}   & \textbf{48.2} & \textbf{35.5}  \\
\midrule
Freeze All        & 83.1       & 30.9 & 11.5     & 46.2 & 32.5  \\
Boundary Training & 83.1       & 32.8 & 19.5     & 47.1 & 33.9  \\
Finetune All      & 83.3       & 34.9 & 19.9     & 46.9 & 33.7 
\end{tabular}
\end{table}

\subsection{Additional results on comparison with current methods}
\rev{
We have presented the comparison between our \MethodName{} and the current SOTA method, TAPADL models \cite{robustfy-attention-tapadl}, in Section~\ref{sec:main-results}. Here, we provide an additional comparison with the ViTCA model \cite{attention-nca}. In ViTCA, tokens interact with each other through local self-attention. We aim to compare our proposed \dynperception{} module with the interaction learning scheme in ViTCA. Results are given in Table~\ref{tab:compare-vitca}. Despite having more parameters, the ViTCA-like scheme performs worse in clean accuracy and is on-par with AdaNCA in robustness. This indicates that it is promising to explore the possibility of incorporating ViTCA into our framework. 
}

\begin{table}[]
\centering
\caption{Comparison with ViTCA. ViTCA performs worse in clean accuracy and is on-par with \MethodName{} in robustness. Note that ViTCA has more parameters than our scheme \dynperception{}. The training setting is the same as in Section~\ref{sec:ablation} in the main paper. }
\label{tab:compare-vitca}
\begin{tabular}{c|cccc}
                                        & \# Params (M) & FLOPS (G) & Clean Acc.($\uparrow$) & Attack Failure Rate ($\uparrow$) \\
                                        \toprule
\rowcolor{gray!20}  \textit{\dynperception{}}  & \textbf{27.94}     & \textbf{4.7}   & \textbf{87.18}      & \textbf{22.35}                   \\
ViTCA, scale=1 & 28.48     & 4.8   & 86.78      & 19.47                   \\
ViTCA, scale=2 & 29.08     & 4.9   & 86.72      & 22.28 
        \end{tabular}
\end{table}

\subsection{Additional results on same model architecture but with more layers ($\star$ models)}
\label{sec:model-size-abl-supp}
Due to the computational cost of RVT \cite{rvt} and FAN \cite{fan} (Table~\ref{tab:train-time}), we do not train them with more layers and thus do not have a $\star$ version of these two models. Instead, we train the smaller version of models on ImageNet100. We choose RVT-small-plus (RVT-S) and FAN-small-hybrid (FAN-S) as two baselines, which is in line with our other experiments on ImageNet100. We add one extra layer to RVT-small-plus and one extra layer to the FAN network in FAN-small-hybrid, resulting in RVT-S$^\star$ and FAN-S$^\star$. We compare it with our \MethodName{}-equipped model where \MethodName{} is after layer 9 in RVT-S and after layer 5 in the FAN-S, aligning with Table~\ref{tab:square-attack}. We use the same AutoAttack as described in Section~\ref{sec:insert-pos-supp} to measure the robustness improvement and attack failure rate. Results are shown in Table~\ref{tab:model-size-abl-supp}. We can observe that the increase in the number of parameters and FLOPS does not account for the improvement in robustness. 
\begin{table}[]
    \caption{Results of different ViTs on ImageNet100. $\star$ means the same model design but with more layers. The improvement \MethodName{} brings does not merely stem from the increase in the number of parameters or FLOPS.\\}
    \label{tab:model-size-abl-supp}
    \centering
    \resizebox{\linewidth}{!}{
    \begin{tabular}{c|cc|c|ccc}
        \toprule
                & Params (M) & FLOPS (G)  & Placement (after) & Clean Acc. & Acc. AutoAttack  & Attack Failure Rate \\
        \midrule
        FAN-S \cite{fan} & 25.8 & 6.7 & / & 87.30                & 18.70               & 21.42      \\
        FAN-S$^\star$ \cite{fan}& 27.7 & 7.1 & / & 87.46                & 19.40               & 22.88      \\
        \rowcolor{gray!20} FAN-S-\MethodName{} & 26.1 & 6.9  & 5  & \textbf{88.22}                & \textbf{27.76}               & \textbf{31.47}       \\
        \midrule
        RVT-S \cite{rvt} & 23.0 &  4.7 & / & 87.18                & 29.80               & 34.18       \\
        RVT-S$^\star$ \cite{rvt} & 24.8 & 5.1  & / &   \textbf{87.52}     &      32.14          &  36.72     \\
        \rowcolor{gray!20} RVT-S-\MethodName{} & 23.3 & 4.9  & 9  & 87.50                & \textbf{36.90}               & \textbf{42.17}       \\
        \bottomrule
    \end{tabular}
    }
\end{table}

\subsection{Additional results on layer similarity structure}
\label{sec:supp-layer-sim-results}
\begin{figure}
    \centering
    \includegraphics[width=\linewidth,trim={80 165 80 110},clip]{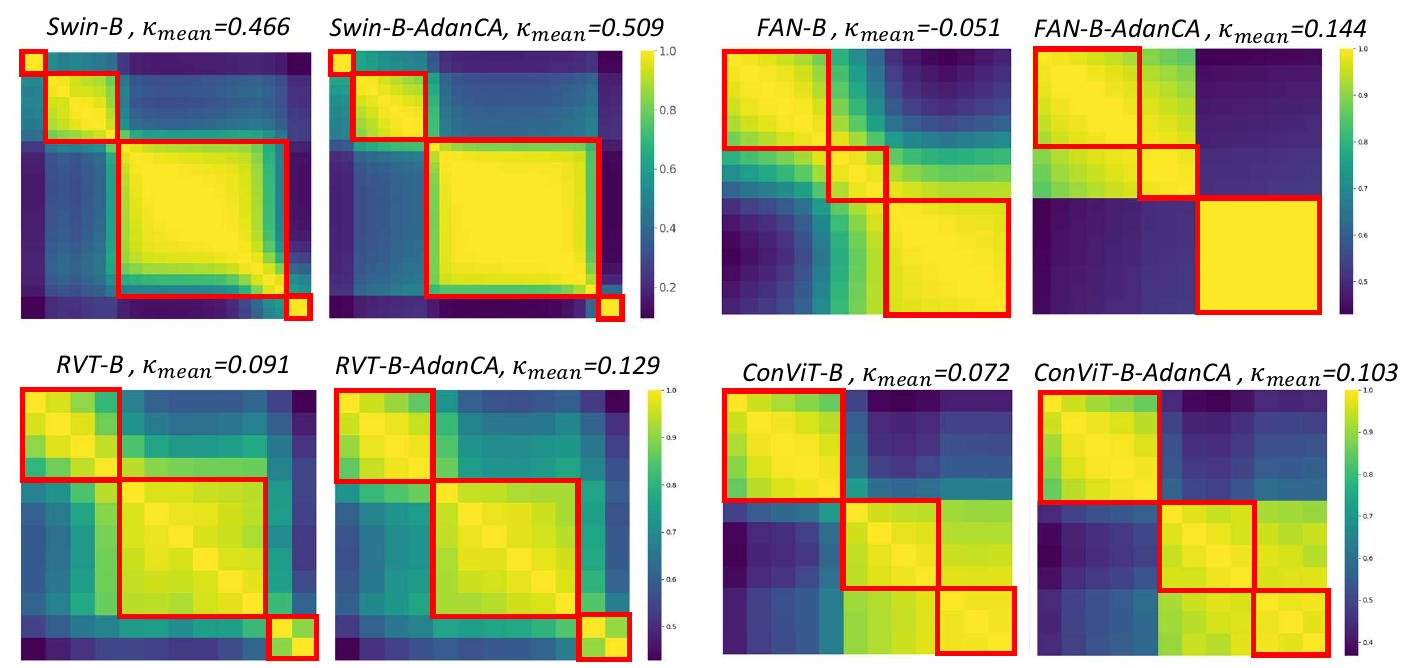}
    \caption{Layer similarity structures of different ViTs. Layer set marked in red boxes. \MethodName{} is inserted between each pair of stages. $\kappa_{mean}$ is the mean of $\kappa$ of all stages marked out.}
    \label{fig:layer-sim-supp}
\end{figure}

Here, we show the layer similarity structure for FAN-B-Hybrid \cite{fan} and ConViT-B \cite{d2021convit} with the ImageNet1K weights and the weights after training with \MethodName{}. For RVT-Base-plus, we train it with the released code \cite{rvt-github} but with a different number of parameters, as discussed in Section~\ref{sec:supp-notes-on-result}. This training is only for obtaining the layer similarity structure, and we find it stable after 30 epochs. The results are shown in Figure~\ref{fig:layer-sim-supp}. In practice, we ignore the stage partition found by the dynamic programming algorithm in Section~\ref{sec:dynamic-programming} that is after the first layer or before the last layer since we want a stage to contain more than 1 layer. The results further validate our assumption that \MethodName{} is used as an adaptor between stages and transmits information between them, making them more and more different from each other. Interestingly, not all \MethodName{} contributes to drastic clearer stage partitions. For example, the first \MethodName{} in FAN-B is not effective in making the stage as clear as the second one. However, the overall Set Cohesion Index increases, and the layer similarity structures in all 4 networks do not change. We believe the deep learning architecture research can benefit from our results, in that developing more architectural changes which contribute to stage partition in the network and examine the effect on network robustness and generalization. More importantly, we use a single non-parametric metric for quantifying the layer similarities. Introducing advanced output similarity quantification methods might lead to new findings and other fascinating results. 

\subsection{Additional results on category-wise mCE on ImageNet-C}
We provide additional results on category-wise mCE on ImageNet-C in Table~\ref{tab:im-c-more-supp}. Our test follows the data transformation used in \cite{rvt}.

\begin{table*}[]
\caption{Comparisons of corruption error on each corruption type of ImageNet-C. \MethodName{} consistently improves the performance of different ViT models.\\}
\label{tab:im-c-more-supp}
\resizebox{\linewidth}{!}{
\begin{tabular}{c|c|ccc|cccc|cccc|cccc}
\toprule
\multirow{2}{*}{Model}  &  \multirow{2}{*}{mCE} & \multicolumn{3}{c|}{\textbf{Noise}}  &  \multicolumn{4}{c|}{\textbf{Blur}} & \multicolumn{4}{c|}{\textbf{Weather}} & \multicolumn{4}{c}{\textbf{Digital}} \\
 & &                                        Gauss & Shot & Impulse & Defocus & Glass & Motion & Zoom & Snow & Frost & Fog  & Bright & Contrast & Elastic & Pixelate & JPEG \\
\midrule
Swin-B \cite{swin} & 54.3 &                 43.8    & 45.7 & 43.8    & 61.6    & 73.2  & 55.5   & 66.5 & 50.0 & 47.4  & 47.9 & 42.1       & 38.7     & 68.8    & 66.4     & 62.6 \\
\rowcolor{gray!20} \textbf{Swin-B-\MethodName{}}      &  \textbf{51.5} &  \textbf{39.5} & \textbf{40.9} & \textbf{39.8} & \textbf{59.5} & \textbf{69.5} & \textbf{55.0} & \textbf{65.6} & \textbf{49.4} & \textbf{47.1} & \textbf{40.8} & \textbf{40.6} & \textbf{37.0} & \textbf{66.3} & \textbf{59.9}  & \textbf{61.1}  \\ 
\midrule
TAPADL-RVT \cite{robustfy-attention-tapadl} & 44.7 & 34.6    & 36.1 & 34.0    & 53.9    & 62.3  & 52.2   & \textbf{60.1} & \textbf{41.9} & 44.9  & 35.8 & 37.4       & 31.9     & 57.1    & 41.1     & 47.2 \\
\rowcolor{gray!20} \textbf{RVT-B-\MethodName{}}      &  \textbf{43.2} &  \textbf{33.9} & \textbf{36.1} & \textbf{33.9} & \textbf{52.7} & \textbf{59.1} & \textbf{47.8} & 60.5 & 42.6 & \textbf{39.0} & \textbf{33.9} & \textbf{36.8} & \textbf{31.7} & \textbf{53.4} & \textbf{40.0}  & \textbf{46.2}  \\ 
\midrule
ConViT-B \cite{d2021convit} & 46.9 &                 36.0    & 37.6 & 35.5    & 56.5    & 64.4  & 51.5   & 65.0 & 42.6 & 40.5  & 39.3 & 39.1       & 38.0     & 61.7    & 43.6     & 51.8 \\
\rowcolor{gray!20} \textbf{ConViT-B-\MethodName{}}      &  \textbf{44.3} &  \textbf{34.7} & \textbf{35.9} & \textbf{34.2} & \textbf{53.1} & \textbf{61.0} & \textbf{48.6} & \textbf{60.4} & \textbf{40.5} & \textbf{40.1} & \textbf{37.2} & \textbf{37.2} & \textbf{33.5} & \textbf{57.1} & \textbf{42.4}  & \textbf{49.4}  \\ 
\midrule
FAN-B \cite{fan} & 46.1 &                 36.5    & 36.8 & 34.7    & 52.8    & 65.9  & 47.7   & 56.9 & 40.6 & 44.4  & 37.9 & 37.4       & 34.3     & 62.8    & 52.8     & 50.6 \\
\rowcolor{gray!20} \textbf{FAN-B-\MethodName{}}      &  \textbf{44.7} &  \textbf{35.5} & \textbf{36.0} & \textbf{33.7} & \textbf{51.9} & \textbf{65.7} & \textbf{45.3} & \textbf{55.6} & \textbf{39.7} & \textbf{43.3} & \textbf{37.3} & \textbf{37.0} & \textbf{33.1} & \textbf{62.0} & \textbf{46.8}  & \textbf{47.9}  \\ 
\bottomrule
\end{tabular}
}
\end{table*}

\subsection{Additional results on noise sensitivity examination}
\label{sec:supp-freq-result}
\begin{figure}
    \centering
    \includegraphics[width=\linewidth, trim={100 170 120 150},clip]{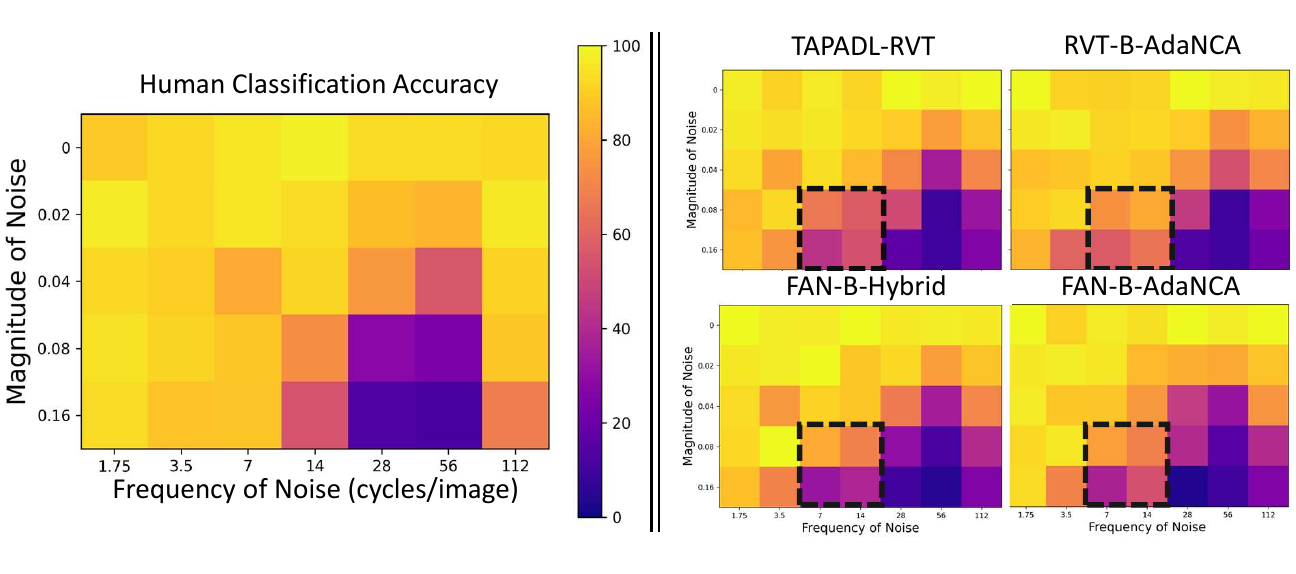}
    \caption{Noise sensitivity examination on humans and ViTs. \MethodName{} contributes to more human-like noise-resilient ability and reduces the sensitivity of ViTs on certain types of noise (yellow boxes).}
    \vspace{-10pt}
    \label{fig:freq-bias-supp}
\end{figure}

We use the data from \cite{freq-preference}. In the experiment, human subjects are asked to classify the noise-contaminated images, and the classification accuracy is recorded. Then, the images are fed into the models trained on ImageNet1K, whose outputs are then transformed into a 16-way ImageNet label \cite{imagenet16} and compared against the ground truth labels. Their classification accuracy is also recorded and used for comparison with human results.
We present the additional results on FAN-B-Hybrid \cite{fan} and the SOTA model TAPADL-RVT \cite{robustfy-attention-tapadl} compared to the \MethodName{}-equipped model in Figure~\ref{fig:freq-bias-supp}. 

Moreover, we adopt a quantitative metric on the test. In the original experiment \cite{freq-preference}, a Gaussian curve is fit based on the discretized accuracy map. However, such a method ignores the continuous change in the classification pattern. Hence, we propose a simple metric that examines the similarity between the accuracy map of humans and of models. Specifically, given the ground truth accuracy map (human performance) $\mathcal{A}_{gt}$, and model accuracy map $\mathcal{A}_m$, the similarity $\Gamma$ is defined as:
\begin{equation}
    \Gamma = 100 - \frac{1}{N} \sum_{\epsilon,f} |\mathcal{A}_{m}(\epsilon,f) - \mathcal{A}_{gt}(\epsilon,f)|
\end{equation}
$\epsilon,f$ represent the magnitude and frequency of the noise, respectively. A higher $\Gamma$ indicates that the model can achieve more similar accuracy maps with humans averaged across all noisy images. $N$ is the total number of types of noise added. Since all models as well as humans perform similarly on low-magnitude noise, we ignore the first two noise levels ($\epsilon=0,0.02$). Hence, in our experiments, $N=21$. We report $\Gamma$ in Table~\ref{tab:freq-quant}. We also include the results of ResNet50 \cite{resnet} as well as the L2-adversarially-trained version \cite{freq-preference} (ResNet50-Adv, trained with L2-bounded adversarial noise where the maximum magnitude of noise is $0.25$), and present the visualizations of the accuracy maps of those two models in Figure. Critically, our results indicate that the adversarially trained model cannot lead to improved accuracy similarities, resulting in a less human-like noise-resilient ability. This is in line with the conclusion of the original work \cite{freq-preference}, validating our proposed metric. Contrary to adversarial training, we obtain more human-like decision patterns than the compared models, validating our claim in Section~\ref{sec:freq-preference}. 

\begin{table}[]
    \caption{Quantitative results on the frequency preference examination experiments. We report the similarity of accuracy maps between model results and humans.\\}
    \label{tab:freq-quant}
    \centering
    \begin{tabular}{c|c}
    \toprule
                                                & Accuracy map similarity to ground truth $\Gamma$ \\
    \midrule
    ResNet50 \cite{resnet} &     72.23      \\
    ResNet50-Adv \cite{freq-preference} & \textcolor{limegreen}{52.12} \\
    \midrule
    \midrule
    TAPADL-RVT \cite{robustfy-attention-tapadl} &     81.91      \\
    RVT-B-\MethodName{} & \textbf{82.17} \\
    \midrule
    FAN-B-Hybrid \cite{fan} & 83.52 \\
    FAN-B-\MethodName{} & \textbf{84.75} \\
    \midrule
    Swin-B \cite{swin}  &  77.07  \\
    Swin-B-\MethodName{} & \textbf{81.34} \\
    \midrule
    ConViT-B \cite{d2021convit} & 79.69 \\
    ConViT-B-\MethodName{} & \textbf{81.56} \\
    \bottomrule
    \end{tabular}
\end{table}

\subsection{Datasets information and license}
\label{sec:dataset-info}
\begin{itemize}[leftmargin=*]
    \item ImageNet1K \cite{deng2009imagenet}. This dataset contains 1.28M training images and 50000 images for validation. We report the top1 accuracy on the 50000 validation images. License: \href{https://image-net.org/}{Custom (research, non-commercial)}.
    \item ImageNet-C \cite{imagenet-c}. This dataset contains 15 types of 2D image corruption types that are generated by different algorithms. The metric on this dataset is mean corruption error, whose lower value represents a more robust model against those corrupted images. License: CC BY 4.0. 
    \item ImageNet-A \cite{imagenet-a}. This dataset contains naturally existing adversarial examples that can drastically decrease the accuracy of ImageNet1K-trained CNNs. It is a 200-class subset of the ImageNet1K dataset. License: MIT license.
    \item ImageNet-R \cite{imagenet-r}. This dataset contains different artistic renditions of 200 classes from the original ImageNet object classes. The original ImageNet dataset discourages non-real-world images, and thus the artistic renditions render the images to be O.O.D. License: MIT license.
    \item ImageNet-SK \cite{imagenet-sk}. This dataset contains 50000 images with 1000 classes that match the validation set of the original ImageNet dataset. All the images are black-and-white sketches instead of real-world photographs of the object class. License: MIT license.
\end{itemize}

\subsection{Model and code license}
\label{sec:model-license}
\begin{itemize}[leftmargin=*]
\item Swin-B \cite{swin}: MIT License.
\item FAN-B \cite{fan}: Nvidia Source Code License-NC.
\item ConViT-B \cite{d2021convit}: Apache 2.0 License.
\item RVT \cite{rvt}: Apache 2.0 License.
\item TAPADL \cite{robustfy-attention-tapadl}: MIT License.
\item Code for adversarial attacks \cite{attack-github}: MIT License.
\item PyTorch Image Model \cite{timm}: Apache 2.0 License.
\item Code for GradCAM++ \cite{jacobgilpytorchcam}: MIT License.
\end{itemize}

\section{Extended ablation studies}
\label{sec:extend-abl}
\rev{In this section, we discuss the design of our \MethodName{} and provide additional ablation studies on the designs of \MethodName{}.}

\subsection{The dropout-like strategy}
\label{sec:dropout-stoch-abl}
\rev{
In Section~\ref{sec:connect-nca-vit}, we introduce our approach of using the dropout-like strategy to perform the stochastic update during training and testing. While previous NCA works claim that the stochasticity can be preserved \cite{mordvintsev2020growing,niklasson2021self-sothtml} or can be simply switched off \cite{attention-nca} during testing, the proposed dropout-like scheme is necessary in our case. First, We highlight that stochasticity during testing can hinder the evaluation of adversarial robustness by producing obfuscated gradients \cite{obfuscated-grad}, leading to the circumvention of adversarial attacks. Table~\ref{tab:stoch-test} illustrates this issue, where we test the classification accuracy under CW attack. Stochasticity also results in inconsistent outputs. This is problematic for practical image classification tasks where reliable output is critical, unlike applications focusing on visual effects \cite{niklasson2021self-sothtml,attention-nca} or collective behaviors \cite{randazzo2020self-classifying}. The change in clean accuracy in Table~\ref{tab:stoch-test} indicates that given the same image, the stochasticity can lead to different decisions, hindering the deployment of the trained models in real-world scenarios. To further demonstrate the necessity of our scheme, we remove the compensation of the output magnitude during training, \textit{i.e.}, we remove the $\frac{1}{p}$ in Equation~\ref{eq:nca-vit-train} during training and directly test the model with a synchronized update. The results are given in Table~\ref{tab:dropout-abl}. The drop in clean accuracy and robustness suggests that downstream ViT layers struggle with varying NCA output magnitudes, indicating the usefulness of our proposed dropout-like method.
}

\begin{table}[]
\centering
\caption{Stochasticity during testing will give a false sense of adversarial robustness. We conduct the same experiments twice, which result in Trials 1 and 2. The training setting is the same as in Section~\ref{sec:ablation} in the main paper. }
\label{tab:stoch-test}
\begin{tabular}{c|cc}
                                 & Clean Acc. ($\uparrow$) & CW Acc. ($\uparrow$) \\
                                 \toprule
\rowcolor{gray!20}  \textit{Test-time Synchronous}            & \textbf{87.18}      & 9.48    \\
\midrule
Test-time Stochasticity, Trial 1 & 87.10      & \textbf{10.72}   \\
Test-time Stochasticity, Trial 2 & 87.06      & 10.60  
\end{tabular}
\end{table}

\begin{table}[]
\centering
\caption{Removing $\frac{1}{p}$ in Equation~\ref{eq:nca-vit-train} during training and directly testing with synchronized update. Our dropout-like compensation scheme improves \MethodName{}'s performance.}
\label{tab:dropout-abl}
\begin{tabular}{c|cc}
         & Clean Acc. ($\uparrow$) & Attack Failure Rate ($\uparrow$) \\
         \toprule
\rowcolor{gray!20}  \textit{Original} & \textbf{87.18}      & \textbf{22.35}                   \\
No $\frac{1}{p}$ & 86.84      & 21.94                  
\end{tabular}
\end{table}

\subsection{\dynperception{}}
\label{sec:dyn-abl-more}
\begin{figure}
    \centering
    \includegraphics[width=0.8\linewidth,trim={150 150 150 120},clip]{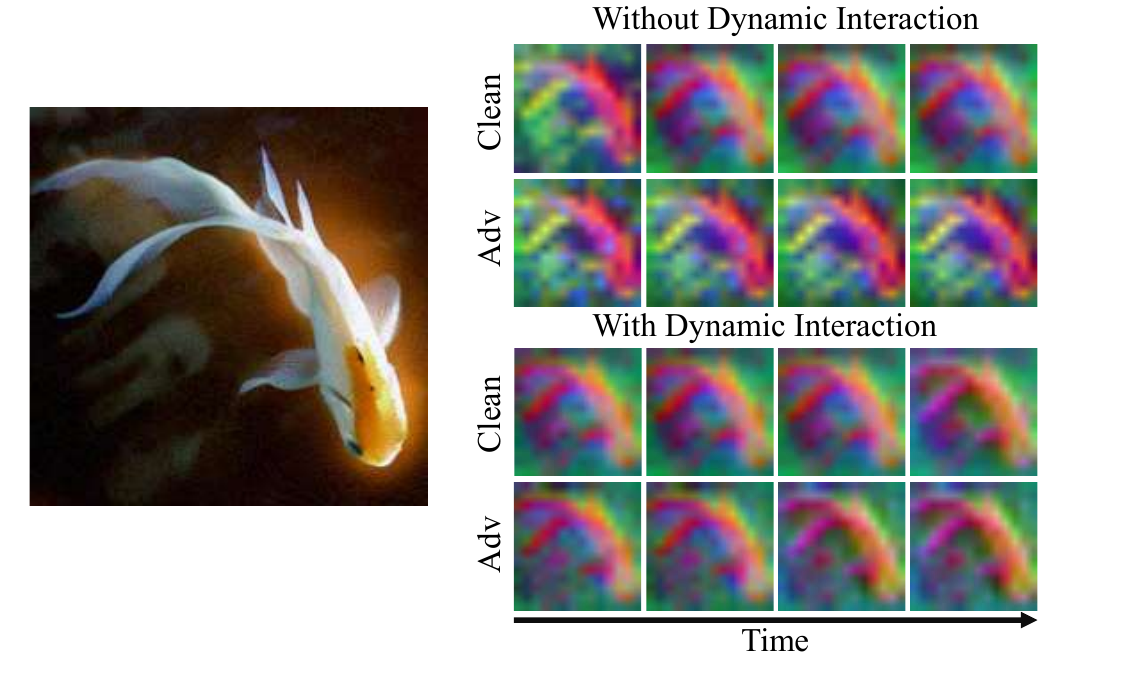}
    \caption{Visualization of the developing token maps of \MethodName{} using PCA. Without \dynperception{}, \MethodName{} cannot recover from the noisy adversarial inputs. Equipped with it, the model can stick to the evolution path similar to the clean inputs.}
    \label{fig:vis-dyn-abl}
\end{figure}

\rev{
Our motivation for developing the \dynperception{} module is the high dimensionality of feature vectors in modern ViT models. We underscore that any operations involving linear transformations of the concatenated interaction results will lead to drastic increases in computational costs, as shown in Table~\ref{tab:concat-param}. Note that the concatenation scheme nearly doubles the FLOPS for RVT compared to the baseline, leading to difficulties in training. Such an increase renders scalability a challenge. Admittedly, more parameters can contribute to better performance, as shown in Table~\ref{tab:concat-abl}. Note, however, that our \dynperception{} achieves ~70\% of the improvements of the original NCA concatenation scheme (10.06/14.62) in robustness improvement and ~60\% improvement in the clean accuracy (0.62/1.06), with merely ~10\% of the parameters and FLOPS (0.35/2.99 for \# Params and 0.2/2.3 for FLOPS). Therefore, our Dynamic Interaction scheme provides a good trade-off between the performance and computational costs. It is capable of scaling up, allowing us to insert \MethodName{} into even larger models, such as current vision-language models where the token dimensionality is even higher. 
Moreover, we qualitatively showcase in Figure~\ref{fig:vis-dyn-abl}, that \dynperception{} helps robustify \MethodName{} when facing noisy inputs.
}

\begin{table}[]
\centering
\caption{The concatenation scheme of the original NCA results in a large increase in the computational costs when the token dimensionality is high, leaving scaling up to large datasets and model sizes a challenge.}
\label{tab:concat-param}
\begin{tabular}{c|cc}
                      & \# Params (M) & FLOPS (G) \\
                      \toprule
Swin-B                & 87.8          & 15.4      \\
\midrule
\rowcolor{gray!20}  \textit{Swin-B-AdaNCA}         & \textbf{90.7}          & \textbf{16.3}      \\
Swin-B-AdaNCA, Concat & \textcolor{red}{115.5}         & \textcolor{red}{24.7}      \\
\midrule
\midrule
RVT-B                 & 88.5          & 17.7      \\
\midrule
\rowcolor{gray!20}  \textit{RVT-B-AdaNCA}          & \textbf{91.0}          & \textbf{19.0}      \\
RVT-B-AdaNCA, Concat  & \textcolor{red}{112.3}         & \textcolor{red}{34.0}    
\end{tabular}
\end{table}

\begin{table}[]
\centering
\caption{Performance of using our Dynamic Interaction and the concatenation scheme of original NCA on ImageNet100. Numbers in parentheses indicate performance improvement compared to the Baseline. Our method achieves a large portion of the performance improvement (60\%~70\%) brought by the concatenation scheme with much fewer computational costs (~10\%).}
\label{tab:concat-abl}
\begin{tabular}{c|cccc}
            & \# Params (M) & FLOPS (G) & Clean Acc.($\uparrow$) & Attack Failure Rate ($\uparrow$) \\
            \toprule
            Baseline     &  27.59  &   4.5  &     86.56   &           12.29 \\
            \midrule
            \rowcolor{gray!20}  \textit{Dynamic Interaction}     &  \textbf{27.94 (+0.35)}  &   \textbf{4.7 (+0.2)}  &     87.18 (+0.62)   &           22.35 (+10.06) \\
            Concat     &  30.58 (+2.99)  &  6.8 (+2.3)  &    \textbf{87.62 (+1.06)}   &          \textbf{26.91 (+14.62)} \\
        \end{tabular}
\end{table}

\subsection{Multi-scale \dynperception{}}
\rev{
Our motivation for developing multi-scale interaction is to perform more efficient token interaction learning over local scales since ViT already contains global information. Enlarging the neighborhood size will: 1) Complicate the process of selecting the neighbors to interact with; 2) Introduce excessive noise, making it difficult for tokens to accurately acquire neighbor information. 3) Repeatedly acquire the global information provided by ViT. The recurrence further amplifies the noise. In our ablation study in Section~\ref{sec:ablation}, we notice that increasing the number of scales to 3 will undermine the performance. Such an issue is also observed in previous work \cite{attention-nca} where local self-attention is used for interaction learning. While in theory both \dynperception{} and self-attention can discount far-away information as the tokens can decide their own interaction neighborhood, overly large neighborhood size can lead to the problems mentioned above.
Note that our model gains performance when $\mathcal{S}=2$ ($5 \times 5$ neighborhood), indicating the usefulness of our multi-scale module. 
We further explore whether the multi-scale issue can be solved by increasing the model capacity. Specifically, we develop two additional schemes. The first is to replace the dilation in \dynperception{} by simply using larger filters. The other one enlarges the receptive field of the weight computation network $\mathcal{W}_{M}$ in Equation~\ref{eq:multi-scale-interaction}. Instead of using two $3\times 3$ convolution layers, we change the first layer to be a $5 \times 5$ convolution. As a result, the receptive field of $\mathcal{W}_{M}$ becomes $7 \times 7$, matching the one when $\mathcal{S}=3$. We use the same training setting as in Section~\ref{sec:ablation}. Results are given in Table~\ref{tab:multi-scale-abl-more}. Although increasing model capacity might alleviate the noise issues, \MethodName{} struggles with overly large neighborhoods.
}
\begin{table}[]
\centering
\caption{Overly large neighborhood size during token interaction will undermine the model performance as it complicates the process of figuring out which neighbors to interact with and adds too much noise during the process. While increasing the model capacity can alleviate such an issue, \MethodName{} still struggles to handle too-noisy information when the scale is too high. Note that the larger filter scheme achieves on-par results with our method when the scale is 2, indicating that it does not provide additional useful information for token interaction at this scale.}
\label{tab:multi-scale-abl-more}
\resizebox{\linewidth}{!}{
\begin{tabular}{c|cccc}
                                   & \# Params (M) & FLOPS (G) & Clean Acc.($\uparrow$) & Attack Failure Rate ($\uparrow$) \\
                                   \toprule
\rowcolor{gray!20} \textit{Original, scale=3}                   & \textbf{27.94}     & \textbf{4.7}   & 86.60      & 20.44                   \\
Larger filter,scale=3              & 28.04     & \textbf{4.7}   & \textbf{86.74}      & \textbf{21.13}                   \\
Larger $\mathcal{W}_M$ receptive field,scale=3 & 27.98     & \textbf{4.7}   & 86.66      & 20.32  \\      
\midrule
\midrule
\rowcolor{gray!20} \textit{Original, scale=2}                   & \textbf{27.94}     & \textbf{4.7}   & \textbf{87.18}      & \textbf{22.35}                   \\
Larger filter,scale=2              & 27.96     & \textbf{4.7}   & 87.14      & 22.31                   \\
\end{tabular}
}
\end{table}

\section{Differences between traditional NCA and \MethodName{}}
\label{sec:diff-nca-adanca}
\rev{
While \MethodName{} is inspired by NCA, it does not fully exploit the designs of traditional NCA. We discuss three key differences between \MethodName{} and NCA below:
\begin{enumerate}
    \item Different initial states. NCA in image generation starts from either constant or randomly initialized cell states, typically referred to as the seed states, while \MethodName{} receives the outputs from previous ViT layers. Hence, \MethodName{} handles structured inputs at the very beginning. 
    \item Different usage of cell states. Traditional NCA does not use all cell states for accomplishing the downstream tasks. After certain steps of evolution, a subset of the cell states is extracted to perform the given task. The unused cell states, termed hidden states, facilitate cell communications as they can be used to store additional information \cite{mordvintsev2020growing}. The dimensionality of the hidden states is typically several times that of the input. When the cell dimensionality is high, as is the case in ViT, adding too many hidden states will bring too much computational costs. Moreover, cells can store effective enough information when they have high dimensionality. Therefore, we do not adopt the hidden states design in \MethodName{}.
    \item No pooling strategy. A critical component of traditional NCA is the pooling strategy. Instead of starting from the seed states in every epoch, NCA fetches the starting states from a pool, where the output states from previous epochs are stored. Through this strategy, NCA can explore much longer time steps without suffering from gradients or memory issues. While it is tempting to incorporate the pooling trick into our method, two major concerns hinder its practical implementation: 1) The previous NCA models start from their own outputs in the pooling strategy. In other words, the cell states in the pool are generated by the model itself. However, \MethodName{} starts from the outputs of a ViT layer. Such a difference renders the pool trick hard to implement. 2) Taking a step back, even if there is a well-designed pooling strategy for \MethodName{}, it will bring too much computational costs during testing. A single step of \MethodName{} introduces non-trivial FLOPS during testing, as shown in Table~\ref{tab:main-result}. The ultimate goal of the pooling strategy is to ensure the stability of NCA in large time steps, while large time steps will slow down the inference. 
\end{enumerate}
}

\end{document}